%% file: main.tex
\documentclass[10pt,twocolumn,letterpaper]{article}

\usepackage[pagenumbers]{cvpr} %

\input{preamble}
\usepackage{enumitem}
\usepackage{subcaption}
\usepackage[ruled,vlined]{algorithm2e}
\providecommand{\sm}{\operatorname{softmax}}
\providecommand{\film}[1]{\emph{#1}}

\def\confName{CVPR}
\def\confYear{2027}

\title{Film Look Transfer}

\author{
Feifan Song$^{1}$ \qquad
Mengyu Ma$^{2}$ \qquad
Yuntian Bo$^{3}$ \qquad
Xiatian Zhu$^{1}$ \\[1ex]
$^{1}$CVSSP, University of Surrey \qquad
$^{2}$Nanjing University of the Arts \qquad
$^{3}$University of Oxford \\[0.5ex]
{\tt\small f.song@surrey.ac.uk, mamengyu@nua.edu.cn, yuntian.bo@eng.ox.ac.uk, xiatian.zhu@surrey.ac.uk}
}
\begin{document}

\twocolumn[{%
\renewcommand\twocolumn[1][]{#1}%
\maketitle
\begin{center}
  \vspace{-3ex}
  \includegraphics[width=\textwidth]{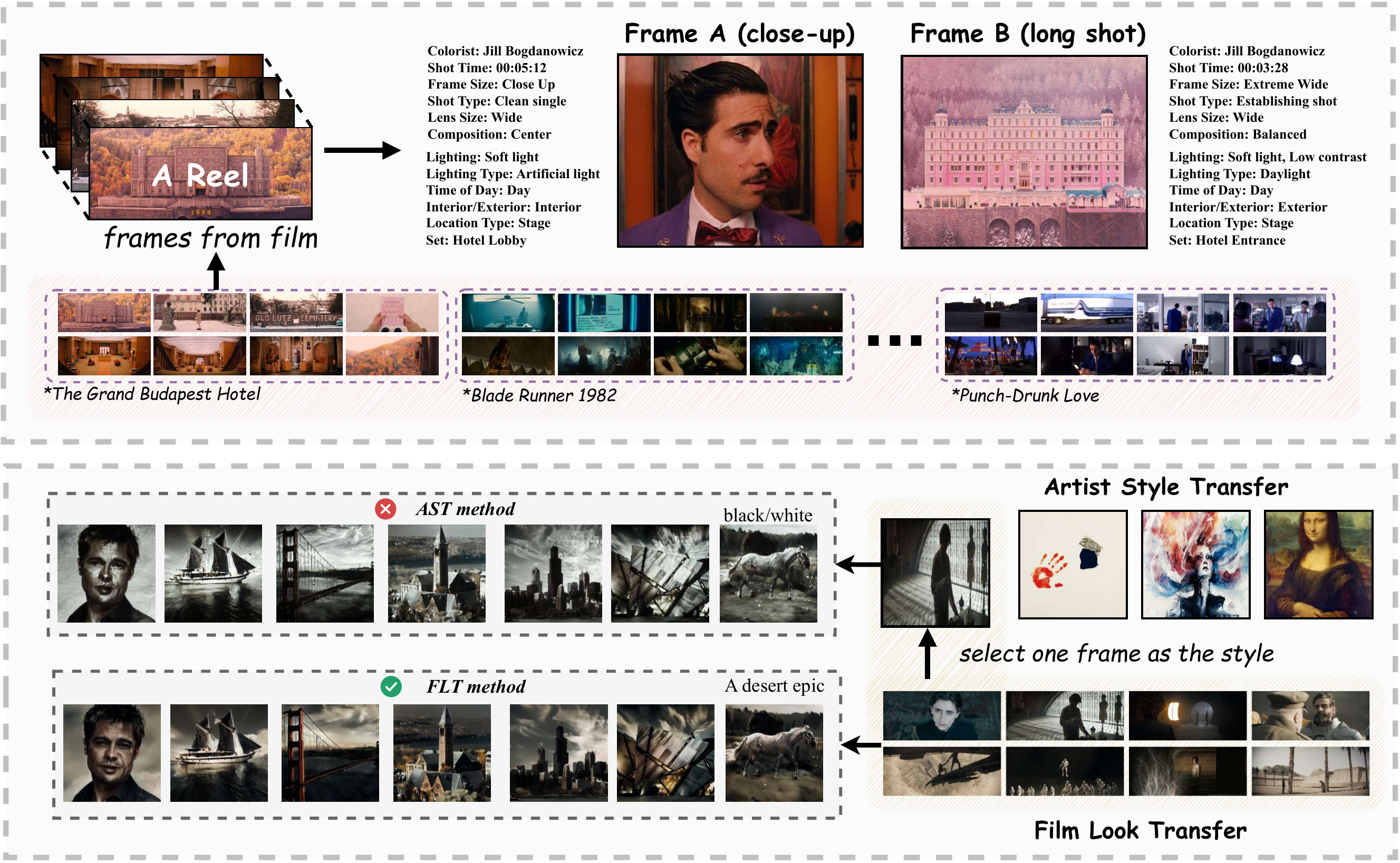}
  \captionof{figure}{
  In the grading suite a film becomes one picture: hundreds of shots, filmed on different days and lenses, are timed into a shared look. \textbf{(Top)} A close-up and an extreme-wide establishing shot from \film{The Grand Budapest Hotel} differ in nearly every mise-en-sc\`ene attribute, \eg, set, lighting and staging, yet both are unmistakably the same \film{Wes Anderson}.
  \textbf{(Bottom)} How to carry such a look onto new footage? Artistic style transfer treats one painting as one style, so it follows whichever frame happens to be chosen: here a single dim interior drives every shot toward black and white. Taking the reel as the reference instead recovers the muted, sand-worn palette of the desert epic \film{Dune}.  
  }
  \label{fig:teaser}
\end{center}
}]

\input{section/abstract}    
\input{section/intro}
\input{section/related}

\input{section/method}

\input{section/benchmark}

\input{section/experiment}

\input{section/conclusion}

\enlargethispage{2\baselineskip}
{
    \small
    \setlength{\bibsep}{0pt plus 0.5pt}  
    \bibliographystyle{ieeenat_fullname}
    \bibliography{references}
}

\input{supp_build}

\end{document}

%% file: preamble.tex
\usepackage[table]{xcolor}
\usepackage{graphicx}
\usepackage{amsmath,amssymb,amsfonts}
\usepackage{booktabs}
\usepackage{multirow}
\usepackage{tabularx}
\usepackage{makecell}
\usepackage{bm}
\usepackage{nicefrac}
\usepackage{microtype}
\usepackage{xr-hyper}
\definecolor{cvprblue}{rgb}{0.21,0.49,0.74}
\usepackage[pagebackref,breaklinks,colorlinks,allcolors=cvprblue]{hyperref}
\usepackage{animate}

\usepackage[table]{xcolor}
\usepackage{booktabs,multirow,tabularx}
\usepackage{capt-of}
\usepackage{wrapfig}

\newif\ifrevmark \revmarktrue

\definecolor{filmcolor}{RGB}{176,62,32}
\newcommand{\film}[1]{\textcolor{filmcolor}{\textit{#1}}}
\definecolor{distcol}{RGB}{176,108,48}
\definecolor{distbg}{RGB}{252,226,196}   

\providecommand{\mF}{{\bm{F}}}
\providecommand{\mM}{{\bm{M}}}
\providecommand{\mS}{{\bm{S}}}
\providecommand{\vd}{{\bm{d}}}
\providecommand{\R}{\mathbb{R}}
\providecommand{\argmin}{\operatorname*{arg\,min}}

\providecommand{\figref}[1]{Fig.~\ref{#1}}
\providecommand{\tabref}[1]{Tab.~\ref{#1}}
\providecommand{\secref}[1]{Sec.~\ref{#1}}
\providecommand{\appref}[1]{Supp.~\ref{#1}}

\providecommand{\cmark}{\ensuremath{\checkmark}}
\providecommand{\xmark}{\ensuremath{\times}}
\providecommand{\pmark}{\ensuremath{\circ}}

%% file: section/abstract.tex
\begin{abstract}
\vspace{-5ex}

A film's look is authored in the grading suite, where a colourist times every shot to a consistent visual signature, \ie, palette, tonality and texture. Matching new footage to that look remains a largely manual, time-consuming process. We study the still-image form of this step, which we term film look transfer (FLT): given a reel, a set of reference frames sampled from one film, transfer the film's shared look to a target frame while preserving its content and composition. 
Existing approaches, from colour transfer to single- and multi-reference diffusion and instruction-based editing, fall short in this task: none reads the look from a reel, carries it beyond colour and keeps the content intact at a practical per-shot cost.
We introduce AnswerPrint, a training-free method built on a frozen diffusion model that follows the colourist's two-pass workflow: a primary grade transfers the global look, and a secondary pass reinstates the content's own attention inside a sparse structural matte, so that fine detail survives the grade. To evaluate FLT, we build Reel-30, a benchmark of thirty films in eight categories with metrics for film identity and local structure. 
On Reel-30, AnswerPrint achieves the best trade-off, and human studies with trained colourists and a video diagnostic show its practical value as an assistive tool for film post-production.
Code and benchmark manifests will be released.

\end{abstract}

%% file: section/intro.tex
\begin{figure}[t]
  \centering
  \includegraphics[width=\linewidth]{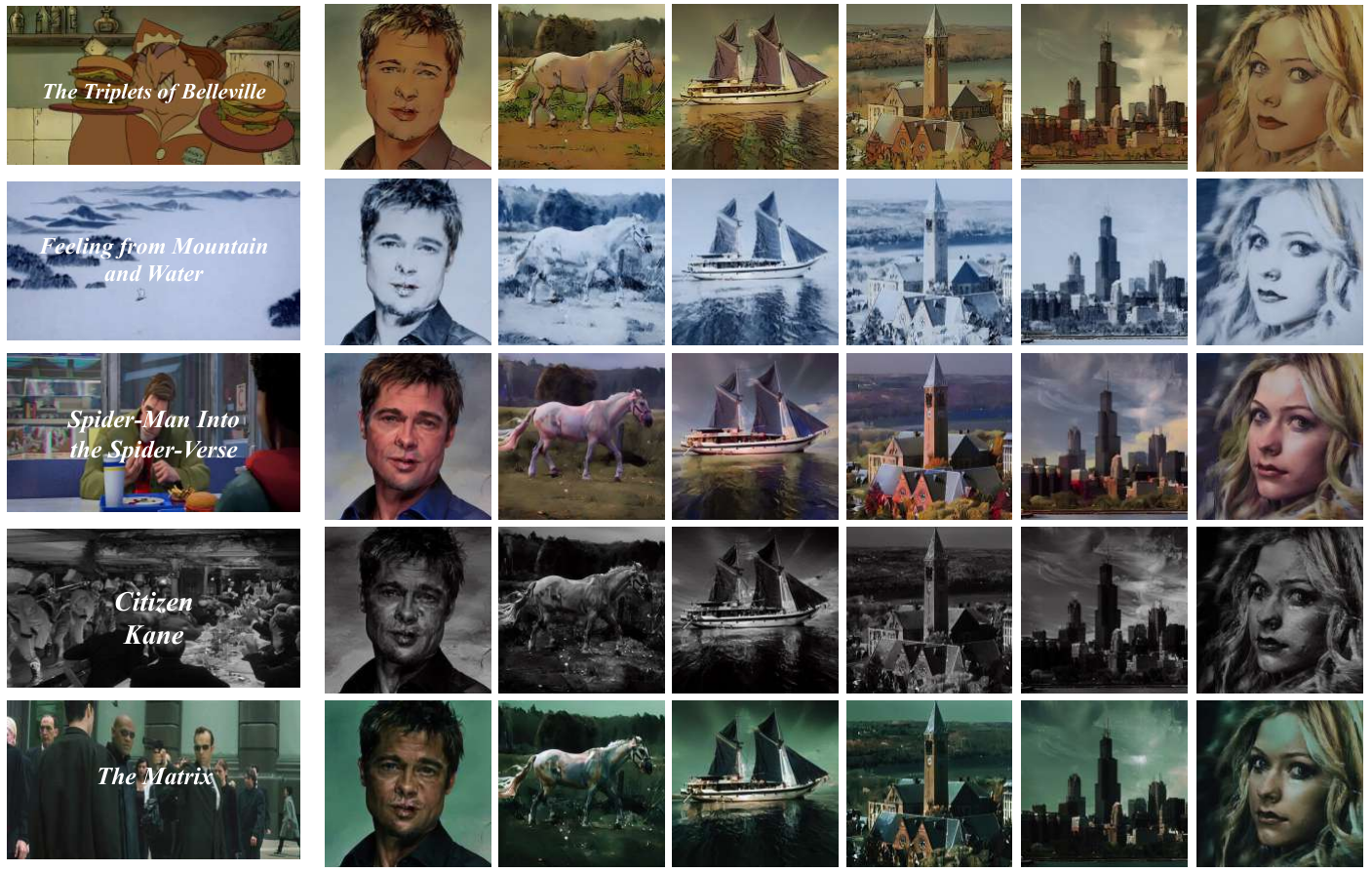}
  \vspace{-4ex}
  \caption{Film look transfer results of \textit{AnswerPrint}.}
  \label{fig:results}
  \vspace{-3ex}
\end{figure}

\vspace{-2ex}
\section{Introduction}
\label{sec:intro}

A film is remembered as much for how it looks as for what it shows: the sun-bleached ochre of \film{Dune}'s Arrakis, the smoky neon blue of \film{Blade Runner}'s rain-soaked Los Angeles, the pastel pinks of \film{The Grand Budapest Hotel}. That look is not captured by the camera. Raw footage arrives flat and inconsistent, shot under different light, lenses and days, and the look is authored afterwards in post-production~\citep{giorgianni2008,vanhurkman2013}, where a colourist times every shot of the film into one visual language. Once the look exists, every new piece of footage that must sit inside it, a reshoot, an inserted VFX plate, a pick-up shot, has to be matched to it by hand, and a single shot can take tens of minutes to bring from an ungraded state to an accepted match.

In this paper, we ask whether a frozen image model can perform this matching step, and study its still-image form as \textbf{\textit{film look transfer} (FLT)}: a \emph{reel}, a set of reference frames sampled from one film, represents its palette, tonality and texture, which is transferred to a target frame while the target's content is kept. As \figref{fig:teaser} illustrates, FLT differs from artistic style transfer (AST), where a single painting defines the style, in that the look is read from the reel --- what persists as its mise-en-sc\`ene changes from shot to shot.

This difference, however, challenges existing tools. Colour transfer, \eg, LUTs, are fast and keep geometry, but move colour alone.
Single-reference diffusion style-transfer methods take one reference image, so the result depends on which frame was picked, whereas multi-reference methods accept a reel but were designed for artistic styles, and pay for it with per-style fine-tuning or per-image optimisation. 
General-purpose image editors were built for open-ended edits rather than grading and run a large model on every shot. Given a reel, they tend to \emph{recreate} the frame instead of \emph{regrading} it, or, defensively, leave it nearly unchanged.
Moreover, none of them has an explicit mechanism that protects the target's fine structure. Analysing FLT as a step of AI-assisted filmmaking, we formalise these challenges as four demands: \textit{\textbf{(R1)} a reel as reference, \textbf{(R2)} a look beyond colour, \textbf{(R3)} intact content and \textbf{(R4)} per-shot practicality}, and find that each existing paradigm misses at least one.

We therefore propose \textit{AnswerPrint}\footnote{Two names borrow from film production: \textit{answer print} is the first fully colour-graded copy against which release prints are matched~\citep{britannica_answerprint}; and \textit{reel}, traditionally a spool of film, to mean a curated set of representative frames.}, 
a training-free FLT method on a frozen diffusion model that follows the colourist's workflow:
a \textit{primary grade} transfers the film's global look, using statistics pooled over the reel and the attention of its medoid frame; a \textit{secondary pass} reinstates the content's own attention inside a sparse structural matte, so that fine detail survives the grade. The reel is inverted once per film, and the whole method runs in under 5\,GB of GPU memory. 
To evaluate FLT, we construct \textit{Reel-30}, a benchmark of 1{,}432 curated frames from thirty films in eight categories, from animation to live action and from monochrome to saturated, with held-out frames and metrics for film identity and local structure. 
We further validate \textit{AnswerPrint} with trained colourists, both against existing methods and as a starting point for their own grading.

In summary, our \textit{contributions} are as follows:
\begin{itemize}[leftmargin=*, itemsep=1pt, topsep=2pt, parsep=0pt]
  \item We formulate FLT as the look-matching step of AI-assisted filmmaking, analysing what the step demands and how existing paradigms meet it.
  \item We propose \textit{AnswerPrint}, a training-free method whose two passes separate the global look from local structural restoration, balancing transfer quality and efficiency.
  \item We build Reel-30, a film-still benchmark, covering eight film categories, from animation to live action, with metrics for film identity and local structure.
  \item We compare against colour-transfer, single- and multi-reference and image-editing baselines, and validate results with human studies and a video diagnostic for practice.
\end{itemize}

%% file: section/related.tex
\vspace{-1ex}
\section{Film Look Transfer: An Analysis}
\label{sec:related}

Grading predates the computer: silent films were tinted and toned by hand, laboratories timed every print by eye, and \film{O Brother, Where Art Thou?} (2000) was the first feature graded entirely digitally~\citep{hullfish2012art}.
AI has long reached into filmmaking: from rule-based virtual cinematography \citep{he1996virtual} to camera-controlled video generation \citep{he2025cameractrl}, it has addressed the layers of film style that concern the camera and the shot.
The look of the film, however, is authored in post-production, and FLT is the step of carrying that look onto new footage. Here, we analyse what the step demands and how existing paradigms meet it.

\vspace{1ex}
\noindent{\bf What FLT demands.} 
This matching step implies four requirements. \textbf{(R1) The reel is the reference:} the look must be read from a set of frames, not from one still that also carries its own scene.
\textbf{(R2) Look beyond colour:} grain, texture and rendering separate films as much as palette does, from the ink wash of \film{Feeling from Mountain and Water} to the silver grain and deep shadows of \film{Citizen Kane} (\figref{fig:results}).
\textbf{(R3) Content is untouchable:} the target's objects stay, no reference content enters, and fine structure survives. \textbf{(R4) Per-shot practicality:} a look is set once and applied to every new shot, so each shot needs no per-film training and little compute. Table~\ref{tab:paradigms} summarises how existing paradigms fare.

\begin{table}[t]
\centering
\caption{\textbf{Paradigms against the demands of FLT}: R1 reel as reference, R2 look beyond colour, R3 content intact, R4 per-shot practicality. \cmark\ met, \pmark\ partly, \xmark\ not met.}
\label{tab:paradigms}
\vspace{-1.5ex}
\small
\setlength{\tabcolsep}{6pt}
\begin{tabular}{lcccc}
\toprule
Paradigm & R1 & R2 & R3 & R4 \\
\midrule
Colour transfer  & \pmark & \xmark & \pmark & \cmark \\
Single-ref.\ diffusion & \xmark & \cmark & \pmark & \cmark \\
Multi-ref.\ diffusion & \cmark & \cmark & \pmark & \pmark \\
Image-editing model & \pmark & \cmark & \xmark & \pmark \\
\rowcolor{gray!20}
\textsc{AnswerPrint} & \cmark & \cmark & \cmark & \cmark \\
\bottomrule
\end{tabular}
\vspace{-4ex}
\end{table}

\noindent {\bf Colour transfer.}
Automating the look began in colour science, with global statistics transfer between an image pair \citep{reinhard2001color}, the closed-form LUT-style grading \citep{pitie2007mkl}, learned colour-histogram control \citep{afifi2021histogan}, image-adaptive and reference-driven 3D LUTs \citep{zeng2020lut, ke2023neural, chen2023nlut}, which recently made spatially adaptive \citep{gong2025salut}. 
These methods are fast and keep geometry, and their statistics can be pooled over a reel. 
However, they stop at colour (R2): grain and rendering style cannot move.

\noindent {\bf Single-reference diffusion style transfer.}
Pre-trained diffusion models move texture as well as tone. StyleID \citep{chung2024styleid} substitutes the reference's self-attention keys and values under the content's queries, StyleAligned \citep{hertz2024stylealigned} shares attention across a batch, InstantStyle \citep{wang2024instantstyle} routes an image-prompt adapter \citep{ye2023ipadapter} into style-specific blocks, and StyleSSP \citep{xu2025stylessp} refines the sampling start point. Embedding-, guidance- and latent-path variants follow the same protocol \citep{zhang2023inst,wang2023stylediffusion,kwon2023diffuseit,jeong2024injectfusion}. Each reads the look from one still (R1), so the result follows whichever frame is chosen (Table~\ref{tab:frame-sensitivity}), and global attention transfer erodes fine structure (R3).

\noindent {\bf Multi-reference diffusion style transfer}.
Recent diffusion work moves from a single reference to a set, the closest existing setting to reading a look from a reel.
\citet{ruta2025multistyle} pool image-prompt embeddings and attention features over a style set, with ControlNet guidance for content. StyleGallery~\citep{he2026stylegallery} matches semantic regions across references and optimises a regional style loss during sampling.
Both satisfy R1 by design, but at the cost of per-style fine-tuning or per-image optimisation (R4). Built for artistic styles, they also have no explicit mechanism for the target's fine structure (R3).

\noindent {\bf Instruction-based image editing.}
General-purpose editors such as Qwen-Image-Edit \citep{wu2025qwenimage}, FLUX.1 Kontext \citep{labs2025kontext} and FLUX.2 \citep{bfl2025flux2} take a source image, reference images and an instruction, and restyle among many other edits. None is designed to read one look from a set of frames, so the reel reaches them as a tiled sheet (R1). Each runs a model of $12$ to $32$ billion parameters per shot (R4), and nothing ties an edit to appearance alone. Given a reel sheet, they either leave the target nearly unchanged or recreate it, their generative strength overwriting the content (R3).

In sum, no existing paradigm meets all four demands. AnswerPrint (\secref{sec:method}) is designed around them: it reads the look from the whole reel (R1), transfers texture with tone (R2), protects fine structure (R3), and runs training-free in under $5$\,GB of GPU memory (R4).

%% file: section/method.tex
\vspace{-1ex}
\section{Method: AnswerPrint}
\label{sec:method}

\subsection{Preliminaries}
\label{sec:prelim}
We build on Stable Diffusion (SD)~\citep{rombach2022ldm}, whose U-Net denoiser interleaves residual, self-attention (SA) and cross-attention (CA) blocks, where SA features fix the layout and appearance of a synthesis. Inversion-based editing exploits this property. Plug-and-Play~\citep{tumanyan2023pnp} injects residual features and SA maps from an inverted source, MasaCtrl~\citep{cao2023masactrl} reuses its SA keys and values, and StyleID~\citep{chung2024styleid} keeps the content's queries while substituting the style's keys and values.
All rely on DDIM inversion~\citep{song2021ddim}: deterministic DDIM sampling maps a start point $z_T$ to an image along a fixed trajectory, and running the update in reverse maps a real image ($t{=}0$) to a $z_T$ that regenerates it, yielding the trajectory $[z_t^{*}]_{t=0}^{T}$ whose per-step SA features can be recorded and re-injected at the same timestep. We use no text prompt throughout, so the CA output is set to zero (\secref{sec:experiments}).

\begin{figure}[t]
  \centering
  \small
\includegraphics[width=\linewidth]{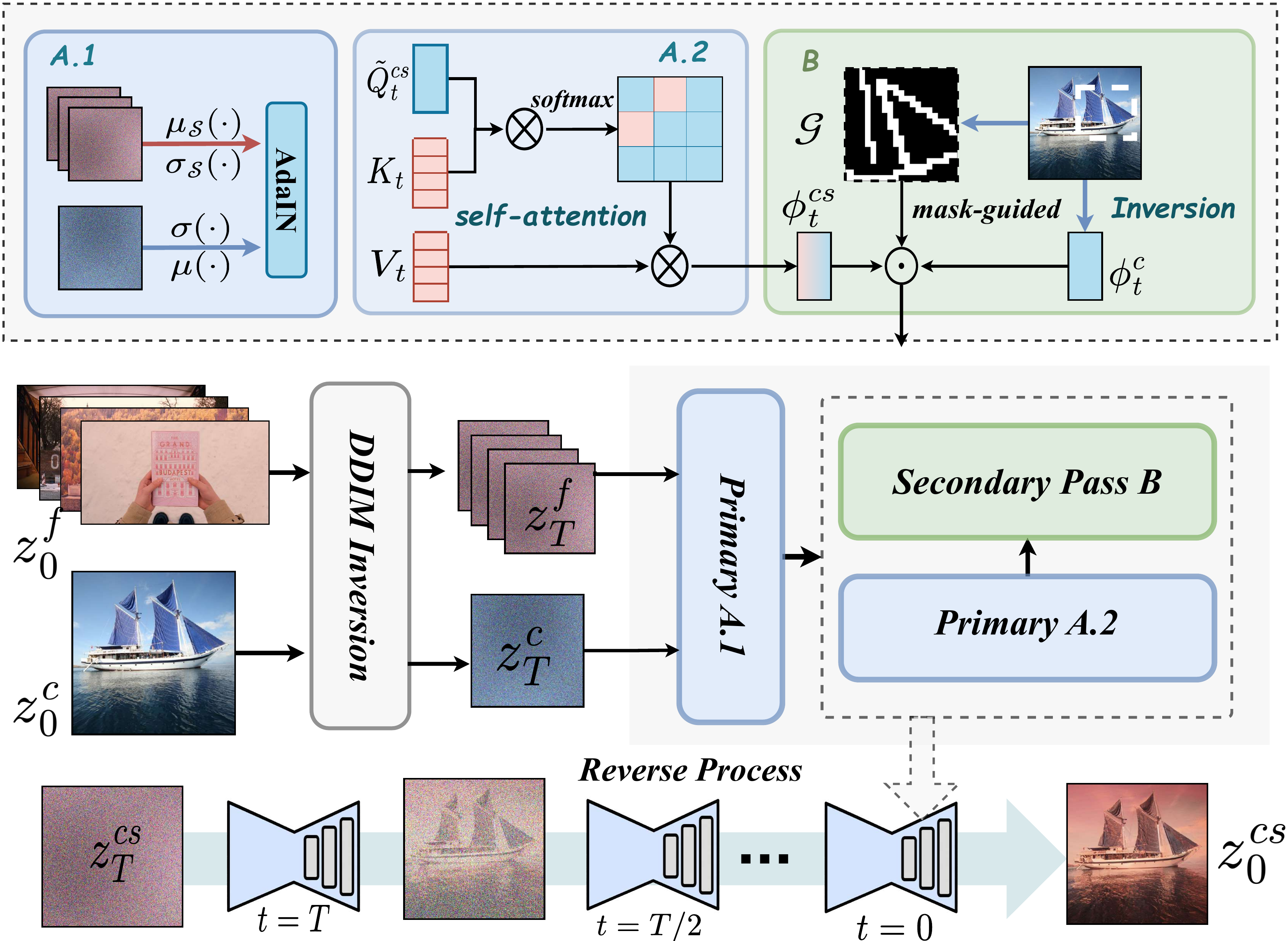}
  \vspace{-3.5ex}
  \caption{
  \textbf{Overview of AnswerPrint.} After DDIM inversion of the reel and the content, the \emph{primary grade} sets the global look through pooled AdaIN initialisation (A.1) and reel key/value injection into self-attention (A.2; \secref{sec:alight}). The \emph{secondary pass} re-instates the content's attention on a sparse structure graph $\mathcal{G}$ (B; \secref{sec:akey}). The backbone stays frozen throughout.  
  }
  \label{fig:pipeline}
  \vspace{-3ex}
\end{figure}

\subsection{Problem Formulation}
\label{sec:overview}
Unlike artistic style transfer, where a single image $I_s$ defines the style, FLT reads the look of a film $\mathcal{S}$ from a reel of $F$ frames $\mathcal{I}_s=\{I_s^{1},\dots,I_s^{F}\}$. Given a content image $I_c$, the goal is a stylized image $I_{cs}$ that keeps the content and composition of $I_c$ while carrying the look shared by $\mathcal{I}_s$. We work in the latent space of SD and denote the encoded content, reel frames and stylized image by $z_0^{c}$, $\{z_0^{f}\}_{f=1}^{F}$ and $z_0^{cs}$, and the DDIM-inverted latents by $z_T^{c}$ and $z_T^{f}$.

As shown in \figref{fig:pipeline}, AnswerPrint follows the colourist's two-pass workflow on a frozen SD backbone. The content and the reel frames are DDIM-inverted once, and their SA features are recorded along the inversion trajectories. Each reverse step then applies two passes in sequence, a primary grade (\secref{sec:alight}) and a secondary pass (\secref{sec:akey}).

\subsection{The Primary Grade}
\label{sec:alight}

The primary grade carries the look shared across the reel onto the content, the diffusion analogue of a colourist's global grade.
During inversion, we record the query features $Q^{c}_{t}$ and SA output $\phi^{c}_{t}$ of the content, and the key and value features $K^{f}_{t},V^{f}_{t}$ of the reel frames, at every timestep $t\in\{0,\dots,T\}$, omitting the layer index for readability.
At each timestep $t$ we assemble the keys and values from $F_b$ banked frames,
\begin{equation}
K_{t}=\big[K_{t}^{1};\cdots;K_{t}^{F_b}\big],\qquad
V_{t}=\big[V_{t}^{1};\cdots;V_{t}^{F_b}\big],
\label{eq:bank}
\end{equation}
where $[\,\cdot\,]$ denotes concatenation and $K_{t},V_{t}\in \mathbb{R}^{(F_b N)\times d}$ stack the $N$ spatial tokens of each frame.
During sampling, we transfer the look to the stylized latent by injecting $K_t$ and $V_t$ into the SA layers, replacing the original $K_t^{cs}$ and $V_t^{cs}$. Following StyleID~\citep{chung2024styleid}, we blend the query of the stylized latent $Q_t^{cs}$ with that of the content $Q_t^c$, $\tilde{Q}_t^{cs}=\gamma\,Q_t^{c}+(1-\gamma)\,Q_t^{cs}$, for the entire reverse process. The style injection at timestep $t$ is then
\begin{equation}
\phi_t^{cs}  \;=\; \operatorname{Attn} \!\left(\tilde{Q}_t^{cs}, K_t, V_t\right).
\label{eq:attn}
\end{equation}
The initial latent carries the look as well, since tone is closely tied to its channel statistics~\citep{chung2024styleid}. We therefore initialise $z_T^{cs}$ from the content noise $z_T^{c}$ by AdaIN~\citep{huang2017adain} against statistics pooled over the inverted reel latents,
\begin{equation}
z_T^{cs} \;=\; \left(\frac{z_T^c-\mu(z_T^{c})}{\sigma(z_T^{c})}\right)\sigma_{\mathcal{S}} + \mu_{\mathcal{S}},
\label{eq:adain}
\end{equation}
where $\mu(\cdot)$ and $\sigma(\cdot)$ denote the channel-wise mean and standard deviation, and $\mu_{\mathcal{S}},\sigma_{\mathcal{S}}$ are the same statistics taken over the set $\{z_T^{f}\}_{f=1}^{F_a}$, with the $F_a$ latents pooled over frames and spatial positions jointly as one population.
The initial latent thus matches the tone of the reel as a whole rather than of any single frame. \secref{sec:where-set-enters} ablates both entry points.

\subsection{The Secondary Pass}
\label{sec:akey}
On top of the indiscriminate primary grade, we deploy a secondary correction to refine object details in the scene. Following the colourist's practice of power windows and qualifiers, this pass is governed by a matte that delimits where the correction takes effect.
The matte has to satisfy two constraints. 
It must be \emph{sparse}, because every anchored token,
\setlength{\intextsep}{3pt}
\begin{wrapfigure}[12]{r}{0.48\linewidth}
    \vspace{-1ex}
    \makebox[\linewidth][r]{\includegraphics[width=1.1\linewidth]{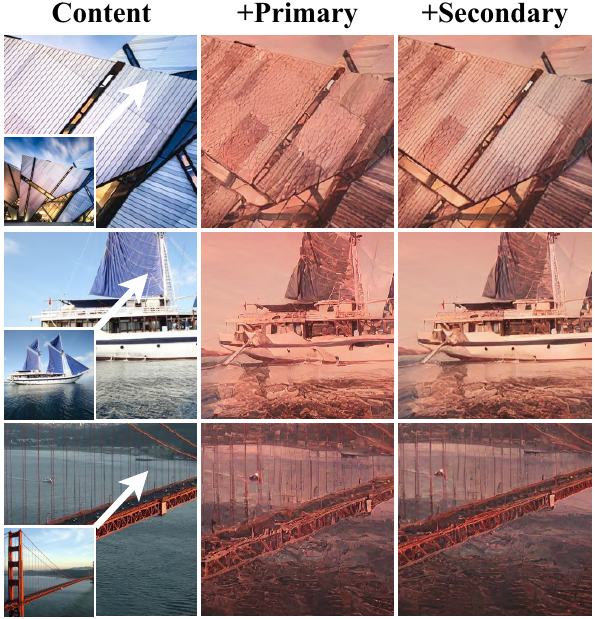}}
    \vspace{-4.5ex}
    \small
    \caption{Effect of the pass.}
    \label{fig:akey}
\end{wrapfigure}
\ie, one covered by the matte, is where the film's look is partly withdrawn, so a dense matte undoes the primary grade. 
And it must lie on \emph{structure} a viewer reads, because restoration spent on texture, such as foliage, ripples or grain, is exactly where the film's own look should prevail.
We therefore realise the matte as a sparse graph $\mathcal{G}$ of the content's long, connected structure, built in three steps (see \appref{app:graph} and Alg.~\ref{alg:anchor}). \textbf{a.} Line segments are detected with LSD~\citep{gioi2008lsd} and merged into continuous polylines by grouping on orientation, collinearity and separation. \textbf{b.} Only corners~\citep{shi1994tomasi} within a snap distance of a retained segment are kept, so that every node rests on an edge rather than floating in texture. \textbf{c.} Connected components whose extent falls below a threshold are discarded as residue. The graph $\mathcal{G}$ is constructed once from the content alone before sampling and reused throughout.

After that, we apply the pass only in the SA layers of the decoder's later, higher-resolution stages, which govern local texture and fine structure, whereas the earlier stages, which set the global look, are left untouched.
For each such layer $l$, we rasterize $\mathcal{G}$ and reduce it to the layer's resolution,
\begin{equation}
M_l[u]\;=\;\max_{p\,\in\,\mathcal{C}_l(u)} R[p],
\label{eq:maxpool}
\end{equation}
where $R$ denotes the binary raster of $\mathcal{G}$ at the content's resolution, $\mathcal{C}_{l}(u)$ is the set of pixels covered by cell $u$ of layer $l$'s grid, and $M_l\in\{0,1\}^{H_l\times W_l}$ is the induced mask.
We adopt a maximum since a one-pixel line occupies only a marginal fraction of a cell and would vanish under averaging. On the masked tokens, we re-instate the content's own recorded attention output,
\begin{equation}
\begin{aligned}
\phi_t^{cs}\;\leftarrow\;\phi_t^{cs}\odot(1-w_aM_l)
\;+\;\phi_t^{c}\odot(w_aM_l),
\end{aligned}
\label{eq:anchor}
\end{equation}
where $\phi_t^{cs}$ is the injected output of \eqref{eq:attn} and $\phi_t^{c}$ is the content's own SA output, recorded during the content inversion before any style is injected. The product $\odot$ is element-wise, with $M_l$ broadcast across channels, and the anchor dose $w_a\in[0,1]$ sets the degree of blending, with a higher $w_a$ drawing the anchored structure closer to the content (\appref{app:dose}).

\begin{figure}[t]
    \centering
    \includegraphics[width=0.9\linewidth]{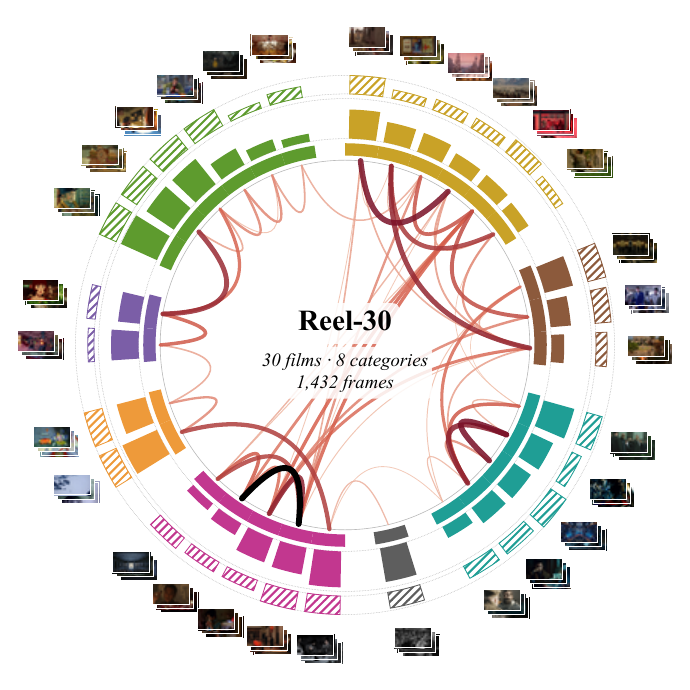}
    \vspace{-3ex}
    \caption{\textbf{Reel-30.} Thirty films (1{,}432 curated frames), shown from the inside out as category, compactness, linear-probe identifiability and each film's three most central frames. Chords link film pairs based on their CSD centroid cosines.}
    \label{fig:reel30}
    \vspace{-3ex}
\end{figure}

%% file: section/benchmark.tex
\vspace{-1ex}
\section{Benchmark: Reel-30}
\label{sec:benchmark}
Grounded in film post-production, we introduce \textbf{Reel-30}, a benchmark for film look transfer in which a look is given by a reel. Existing style-transfer benchmarks define a style as a single artistic image and score appearance with whole-image statistics that are blind to local structure a viewer reads.

\noindent {\bf Films, Reels and Plates.}
\label{sec:reel-construction}
\label{sec:reel-stats}
Reel-30 comprises 1{,}432 curated frames from 30 films in eight categories (\figref{fig:reel30}). The films span 20 live-action and 10 animated titles across eight decades, from \emph{Citizen Kane} (1941) to \emph{Spider-Man: Into the Spider-Verse} (2018), and across American, European, Chinese and Japanese cinema. The animated titles range from ink-wash and oil-painted to stop-motion and computer animation, and each film carries a pronounced visual signature.
For each film, $F$ frames are drawn deterministically as the reel $\mathcal{I}_s$, and the remaining frames are held out to train and test the film-identity probe $\mathcal{P}$. Every reel is paired with the same 20 MS-COCO~\citep{lin2014coco} content images used by the standard style-transfer protocol~\citep{chung2024styleid,xu2025stylessp}.

A reel can serve as a style reference only if it exhibits both intra-film consistency and inter-film distinctiveness, and Reel-30 satisfies both in the CSD style embedding~\citep{somepalli2024csd}. a. Each film is internally consistent. The mean cosine between a film's curated frames and its centroid ranges from $0.520$ to $0.794$ (mean $0.650$). b. Films are mutually distinct. No two film centroids exceed a cosine of $\tau^\star{=}0.611$ (Table~\ref{tab:reel30-redundant}), and $\mathcal{P}$ identifies every film with a recall of at least $0.90$, reaching $0.987$ accuracy on held-out frames.

\vspace{1ex}
\noindent{\bf Evaluation metrics.}
\label{sec:reel-metrics}
We evaluate FLT from three complementary aspects:
\begin{enumerate}[label=(\roman*), leftmargin=1.5em, itemsep=1pt, topsep=2pt, parsep=0pt]

\item \emph{\textbf{ArtFID family}}: the standard style-transfer protocol, $\text{ArtFID}=(1+\text{LPIPS})\cdot(1+\text{FID})$~\citep{wright2022artfid}, where LPIPS~\citep{zhang2018lpips} measures the perceptual distance between each output and its content image, and FID~\citep{heusel2017fid} the distance between the distributions of outputs and style images.

\item \emph{\textbf{Film identity}}: measuring how well an output keeps the visual character of its source film. \textbf{Film Top-1} is the rate at which $\mathcal{P}$ recognises an output as its source film~\citep{sanakoyeu2018deception}, read against the probe's ceiling of $0.987$. \textbf{Style similarity} measures feature-space closeness to the reference frames, via the DINO/CLIP-I protocol of DreamBooth \citep{ruiz2023dreambooth} with DINOv2 features \citep{oquab2023dinov2}.

\item \emph{\textbf{Local structure}}: measuring whether fine content structure survives the grade, which the whole-image metrics in (i) and (ii) cannot reveal. We report \textbf{CFSD}~\citep{chung2024styleid} and \textbf{edge $\boldsymbol{F_1}$} under the condition-reconstruction protocol of \cite{li2024controlnetpp}: Canny edges \citep{canny1986edge} ($3{\times}3$ blur, thresholds $100/200$) are matched between output and content within $2$ pixels.

\end{enumerate}

%% file: section/experiment.tex
\begin{table*}[t]
\small
\centering
\caption{\textbf{Quantitative comparison on Reel-30} (30 films $\times$ 20 contents, 600 outputs per method). Every cell is shaded by its rank among all methods in that column, from purple (best) through white (median) to grey (worst). The \textbf{\emph{balance}} bar shows the gap between a method's mean rank on film identity and on structure and the side it leans to.
}
\label{tab:main}
\vspace{-1.5ex}
\setlength{\tabcolsep}{3pt}
\newcolumntype{C}{>{\centering\arraybackslash}p{46pt}}
\definecolor{hmbest}{RGB}{178,162,230}
\definecolor{hmworst}{RGB}{190,197,206}
\definecolor{barcol}{RGB}{120,105,170}
\newcommand{\wk}[2]{\cellcolor{hmworst!#1!white}$#2$}
\newcommand{\gbar}[2]{\makebox[1.6cm][c]{\makebox[0.8cm][r]{\textcolor{barcol}{\rule{#1cm}{1.1ex}}}\makebox[0pt]{\rule[-0.3ex]{0.4pt}{1.7ex}}\makebox[0.8cm][l]{\textcolor{barcol}{\rule{#2cm}{1.1ex}}}}}
\newcommand{\rk}[2]{\cellcolor{hmbest!#1!white}$#2$}

\resizebox{\textwidth}{!}{%
\begin{tabular}{@{}ll CCCCCCC @{\hspace{8pt}} c@{}}
\toprule
 & & \multicolumn{2}{c}{\textbf{Film identity}} & \multicolumn{2}{c}{\textbf{Structure}} & \multicolumn{3}{c}{\textbf{ArtFID family}} & \textbf{Balance} \\
\cmidrule(lr){3-4} \cmidrule(lr){5-6} \cmidrule(lr){7-9} \cmidrule(lr){10-10}
Method & Backbone & Film top-1 $\uparrow$ & Style sim.\ $\uparrow$ & Edge $F_1$ $\uparrow$ & CFSD $\downarrow$ & FID $\downarrow$ & LPIPS $\downarrow$ & ArtFID $\downarrow$ & {\scriptsize structure\,$\leftarrow$\,$\rightarrow$\,look} \\
\midrule
\multicolumn{10}{l}{\textit{Colour transfer}} \\
\midrule
MKL~\cite{pitie2007mkl} & -- & \wk{22}{0.187} & \wk{11}{0.085} & \rk{67}{0.812} & \rk{78}{0.093} & \rk{56}{19.17} & \rk{67}{0.263} & \rk{78}{25.48} & \gbar{0.414}{0} \\
IDT~\citep{pitie2007automated} & -- & \wk{33}{0.153} & \wk{22}{0.082} & \rk{89}{0.858} & \rk{100}{0.038} & \wk{67}{20.64} & \rk{78}{0.248} & \rk{67}{27.01} & \gbar{0.569}{0} \\
ReHistoGAN~\citep{afifi2021histogan} & GAN & \wk{67}{0.112} & \wk{44}{0.063} & \rk{100}{0.972} & \rk{89}{0.043} & \wk{33}{20.28} & \rk{100}{0.141} & \rk{100}{24.27} & \gbar{0.698}{0} \\
SA-LUT$^{\triangle}$~\citep{gong2025salut} & ViT & \wk{56}{0.117} & \wk{56}{0.059} & \rk{78}{0.841} & \rk{56}{0.124} & \wk{56}{20.57} & \rk{89}{0.173} & \rk{89}{25.31} & \gbar{0.569}{0} \\
\midrule
\multicolumn{10}{l}{\textit{Single-reference style transfer$^{\triangle}$}} \\
\midrule
PnP$^{\ddagger}$~\cite{tumanyan2023pnp} & SD1.5 & \wk{89}{0.075} & \wk{78}{0.026} & \rk{22}{0.638} & \wk{11}{0.201} & \rk{67}{19.12} & \rk{33}{0.380} & \rk{56}{27.76} & \gbar{0.414}{0} \\
StyleAligned~\cite{hertz2024stylealigned} & SDXL & \wk{44}{0.133} & \wk{67}{0.048} & \wk{78}{0.411} & \wk{56}{0.274} & \rk{100}{17.92} & \wk{67}{0.668} & \wk{44}{31.55} & \gbar{0}{0.052} \\
InstantStyle~\cite{wang2024instantstyle} & SDXL & \rk{33}{0.525} & \rk{44}{0.211} & \wk{67}{0.413} & \wk{33}{0.222} & \wk{44}{20.35} & \wk{56}{0.594} & \wk{67}{34.03} & \gbar{0}{0.414} \\
InjectFusion~\cite{jeong2024injectfusion} & ADM & \rk{67}{0.587} & \rk{67}{0.218} & \wk{100}{0.107} & \wk{78}{0.348} & \rk{22}{19.65} & \wk{100}{0.888} & \wk{100}{38.99} & \gbar{0}{0.724} \\
DiffuseIT~\cite{kwon2023diffuseit} & ADM & \rk{11}{0.337} & $0.155$ & \wk{89}{0.338} & \wk{67}{0.278} & \wk{78}{20.66} & \wk{78}{0.676} & \wk{89}{36.30} & \gbar{0}{0.388} \\
StyleID~\cite{chung2024styleid} & SD1.5 & \rk{56}{0.563} & \rk{56}{0.213} & $0.582$ & $0.183$ & \rk{44}{19.38} & \wk{33}{0.504} & \rk{11}{30.65} & \gbar{0}{0.259} \\
StyleSSP~\cite{xu2025stylessp} & SDXL & \rk{22}{0.350} & \rk{22}{0.164} & \wk{56}{0.417} & \wk{44}{0.229} & \rk{11}{19.94} & \wk{22}{0.492} & \wk{22}{31.25} & \gbar{0}{0.336} \\
\midrule
\multicolumn{10}{l}{\textit{Multi-reference style transfer}} \\
\midrule
StyleGallery~\cite{he2026stylegallery} & SD1.5 & \rk{11}{0.337} & \rk{11}{0.160} & \rk{44}{0.729} & \rk{44}{0.155} & \wk{22}{20.11} & \rk{44}{0.333} & \rk{44}{28.16} & \gbar{0.155}{0} \\
Ruta \etal$^{\S}$~\cite{ruta2025multistyle} & SD1.5 & \rk{100}{0.635} & \rk{100}{0.243} & \wk{33}{0.491} & \wk{89}{0.579} & \rk{89}{18.38} & \wk{89}{0.713} & \wk{56}{33.19} & \gbar{0}{0.750} \\
StyleID + pooled$^{\dagger}$ & SD1.5 & \rk{89}{0.613} & \rk{89}{0.227} & \rk{11}{0.606} & \rk{11}{0.181} & $20.08$ & \wk{11}{0.483} & \wk{33}{31.26} & \gbar{0}{0.362} \\
\midrule
\multicolumn{10}{l}{\textit{General-purpose image editing$^{\star}$}} \\
\midrule
Qwen-Image-Edit~\cite{wu2025qwenimage} & MMDiT & \wk{11}{0.188} & \wk{33}{0.073} & \wk{22}{0.509} & \wk{22}{0.215} & \rk{33}{19.60} & \wk{44}{0.508} & \wk{11}{31.07} & \gbar{0}{0.000} \\
FLUX.1 Kontext~\cite{labs2025kontext} & DiT & \wk{78}{0.083} & \wk{89}{0.016} & \wk{44}{0.470} & \wk{100}{0.706} & \wk{100}{22.08} & $0.482$ & \wk{78}{34.20} & \gbar{0.052}{0} \\
FLUX.2~\cite{bfl2025flux2} & DiT & \wk{100}{0.060} & \wk{100}{0.008} & \rk{56}{0.778} & \rk{67}{0.104} & \wk{89}{21.32} & \rk{56}{0.313} & \rk{22}{29.30} & \gbar{0.750}{0} \\
\midrule
\multicolumn{10}{l}{\textit{Ours}} \\
\midrule
\textsc{APrint} & SD1.5 & \rk{78}{0.607} & \rk{78}{0.222} & \rk{33}{0.647} & \rk{33}{0.169} & \wk{11}{20.10} & \rk{22}{0.458} & $30.77$ & \gbar{0}{0.207} \\
\textsc{APrint} & SDXL & \rk{44}{0.535} & \rk{33}{0.197} & \wk{11}{0.552} & \rk{22}{0.179} & \rk{78}{18.44} & \rk{11}{0.466} & \rk{33}{28.50} & \gbar{0}{0.155} \\
\bottomrule
\end{tabular}}
\\[1pt]
\begin{minipage}{\textwidth}
\footnotesize
Single-reference methods receive the reel's medoid frame and multi-reference methods the same $F_a$ frames as APrint (details in \appref{app:baselines}).
$^{\triangle}$ Single reference.
$^{\ddagger}$ Text-conditioned (\appref{app:text}).
$^{\star}$ Reel tiled into one sheet.
$^{\S}$ Our reimplementation.
$^{\dagger}$ Our primary grade based on StyleID (\tabref{tab:design}).
\end{minipage}
\end{table*}

\vspace{-1ex}
\section{AnswerPrint vs. Prior Methods}
\label{sec:experiments}

\noindent {\bf Implementation Details.} We conducted all experiments with a frozen Stable Diffusion backbone and adopt DDIM inversion and sampling with a total of $50$ timesteps ($T=50$). Unless mentioned separately, we set anchor dose $w_a=0.5$, $F_a=F=8$, $F_b=1$, where the $F_b$ frame is the medoid of the $F$ reel frames.
Since FLT is purely visual, the prompt is empty throughout. In this regime the cross-attention keys and values are formed from fixed padding embeddings, and its output is nearly identical across contents. We therefore set the cross-attention output to zero during sampling, which removes a content-independent offset at no cost (App.~\ref{app:Clean~Gate}).

\vspace{1ex}
\noindent {\bf Competitors.} We compare four groups of methods from Sec.~\ref{sec:related}:
\emph{\textbf{a. Colour transfer}}: MKL~\citep{pitie2007mkl}, IDT~\citep{pitie2007automated} and ReHistoGAN~\citep{afifi2021histogan}, fitted to colours pooled over the reel, and SA-LUT~\citep{gong2025salut}, which takes one reference frame.
\emph{\textbf{b. Single-reference}} methods receive the reel's medoid frame: PnP~\citep{tumanyan2023pnp}, InjectFusion~\citep{jeong2024injectfusion}, DiffuseIT~\citep{kwon2023diffuseit}, StyleAligned~\citep{hertz2024stylealigned}, InstantStyle~\citep{wang2024instantstyle}, StyleID~\citep{chung2024styleid} and StyleSSP~\citep{xu2025stylessp}.
\emph{\textbf{c. Multi-reference}} methods receive all $F_a$ frames: StyleGallery~\citep{he2026stylegallery}, \citet{ruta2025multistyle}, and StyleID~\citep{chung2024styleid} given the reel through pooled statistics.
\emph{\textbf{d. Image editors}} receive the same $F_a$ frames tiled into one sheet: Qwen-Image-Edit~\citep{wu2025qwenimage}, FLUX.1 Kontext [dev]~\citep{labs2025kontext} and FLUX.2 [dev]~\citep{bfl2025flux2}, each with one fixed instruction that names no film.
We use public implementations at their recommended settings, adaptations listed in \appref{app:baselines}.

\begin{figure*}[t]
  \centering
  \includegraphics[width=\linewidth]{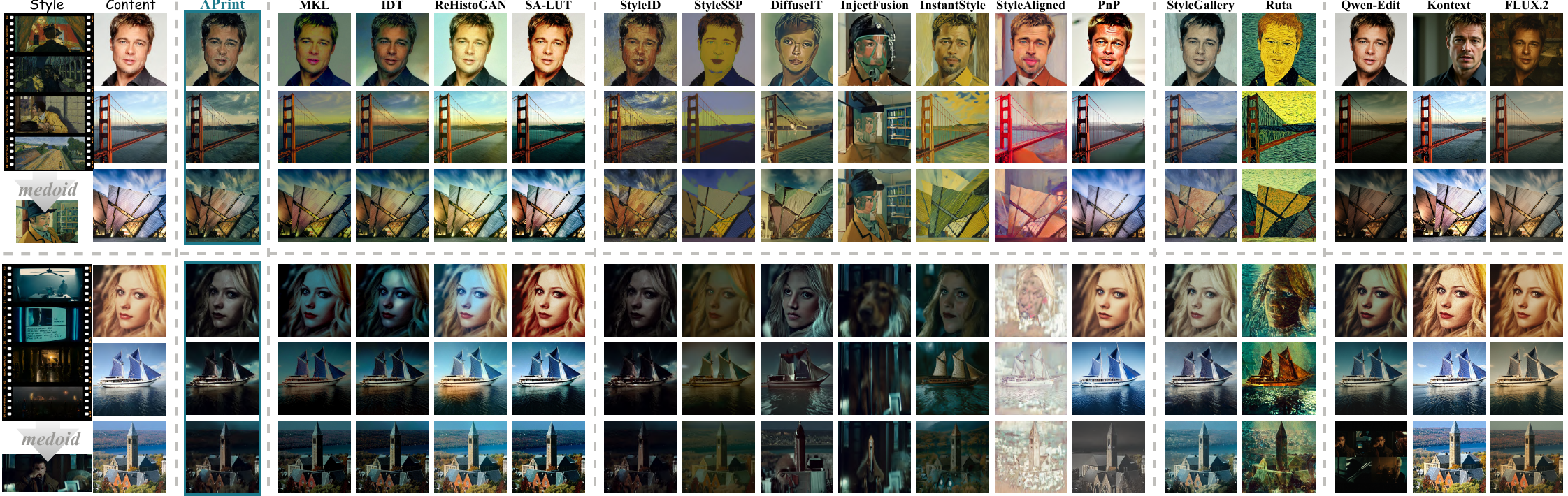}
  \vspace{-4ex}
  \caption{Qualitative comparison with colour-transfer, diffusion and image-editing baselines. \textit{\textbf{Zoom in for better visualisation.}}}
  \label{fig:sota}
  \vspace{-3ex}
\end{figure*}

\subsection{Main Results}
\label{sec:exp-main}
\vspace{-1ex}
Table~\ref{tab:main} and \figref{fig:sota} report quantitative and qualitative results on Reel-30. Film identity and structure pull against each other, and most methods sit at one end of this trade-off. Colour transfer and FLUX.2 preserve structure best, with an edge $F_1$ of $0.778$ to $0.972$, but they barely move the look. At the other end, \citeauthor{ruta2025multistyle} and InjectFusion lead on film identity ($0.635$ and $0.587$) and fall to the bottom on structure. AnswerPrint (APrint) holds both sides. 
It attains a film Top-1 of $0.607$ and an edge $F_1$ of $0.647$, a relative $7.8\%$ above its base model StyleID in film identity. It comes within $0.028$ of the strongest multi-reference method, \citeauthor{ruta2025multistyle}, on film identity while preserving markedly more structure ($0.647$ vs.\ $0.491$; significance tests in \appref{app:results}). Among the methods above $0.5$ film Top-1, APrint is best on edge $F_1$,  CFSD and LPIPS. Visually, APrint carries the reel's palette, tonality and grain onto the plate while faces and fine geometry stay intact, whereas colour transfer methods, StyleGallery, and FLUX.2 return lightly tinted copies of the target, with close to chance on film Top-1.  Powerful generative models, \eg, Qwen-Image-Edit, sometimes overinstate and recreate the raw footage. 
A more detailed analysis is provided in \appref{app:results}.

\noindent \textit{\textbf{The ArtFID family}} shows the same trade-off from the standard protocol. The lowest LPIPS belongs to methods that change the plate least (ReHistoGAN $0.141$, SA-LUT $0.173$), and the lowest FID to methods that depart from the raw footage most (StyleAligned $17.92$ and \citeauthor{ruta2025multistyle} $18.38$). APrint again sits between the two ends, and its SDXL variant reaches the third-lowest FID ($18.44$) at an LPIPS of $0.466$. We therefore assess FLT through film identity and local structure jointly.

\vspace{-1ex}
\section{AnswerPrint with Human}
\label{sec:exp-human}

Film post-production is ultimately a human craft, so we validate APrint with the people it is meant to assist, in collaboration with a professional film academy.

\vspace{1ex}
\noindent {\bf Blind preference.} Eight film practitioners (median $2$ years in the field) each judged the same $120$ blind two-alternative trials: $24$ film--content pairs stratified by category, each pairing APrint with one of ten baselines that span all method groups. Raters saw four held-out frames of the film and the raw photograph, and chose separately the output that \textit{a. better matches the film's look; b. better keeps the photograph's structure} (protocol in \appref{app:human}). 

\noindent Over $960$ judgments APrint is preferred on look in $64.7\%$ (95\% CI $[57.8, 71.5]$, resampling raters and items), clearly over the generative baselines that carry the film (InjectFusion $92\%$, Ruta \etal\ $81\%$, Qwen-Image-Edit $77\%$, FLUX.2 $75\%$, StyleID $68\%$) and level with the rest ($46$--$60\%$). Structure judgments follow edge $F_1$, where raters prefer APrint over the methods that redraw the plate (InjectFusion $99\%$, Ruta \etal\ $95\%$, StyleID $72\%$, InstantStyle $66\%$), but prefer MKL, IDT, StyleGallery and FLUX.2, which leave it almost unchanged ($21$--$33\%$ for APrint), confirming the trade-off of \tabref{tab:main}. 
Per trial, edge $F_1$ agrees with the raters' majority on $75\%$ of structure trials (Cohen's $\kappa{=}0.49$; always choosing APrint agrees on $55\%$), whereas the film probe agrees on $65\%$ of look trials ($\kappa{=}0.16$, against $71\%$ for always-APrint): raters also weigh whether the grade is usable and reject identity-rich outputs that break the plate (InjectFusion, Ruta \etal), so we read film Top-1 as identity rather than overall grade quality. Per-baseline results are in \tabref{tab:human}.

\begin{figure}[t]
  \centering
  \includegraphics[width=\linewidth]{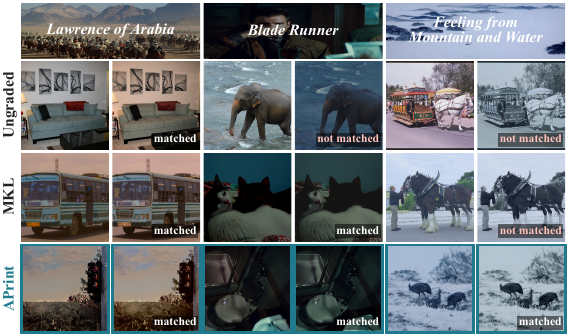}
  \vspace{-4ex}
  \caption{Grading from three starts: the start the colourist received (left of each pair) and the grade reached from it (right), marked \emph{matched} when a second colourist accepted it (\appref{app:human}).}
  \label{fig:lut}
  \vspace{-2ex}
\end{figure}

\vspace{1ex}
\noindent {\bf Manual grading.} One professional colourist graded $8$ Reel-30 films, one per category, in \emph{DaVinci Resolve} from three undisclosed starts, the \textbf{ungraded plate}, \textbf{MKL}~\citep{pitie2007mkl} and \textbf{APrint}, each on a different content. The colourist stopped once the plate matched the reel to their standard, a second colourist, who never saw the starting plate, confirmed the match.
Every APrint start was graded to an accepted match, against $6$ of $8$ from MKL and $4$ of $8$ from the plate. The manual grade can only tint the photograph toward the reel's palette in details, while the tonality and texture that APrint brings in survives. APrint starts also needed the least correction: the colourist's grade changed them by a median of $8.6$ grey levels per pixel, against $12.3$ for MKL and $18.7$ for the plate. In this work, we emphasize that APrint is not the colourist, rather an auxiliary tool, which leaves a lighter final match to them.

\begin{table}[t]
\centering
\caption{\textbf{Component analysis.} The secondary pass and the cross-attention skip are independent additions to the primary grade, and \textsc{AnswerPrint} uses both. We sweep the anchor dose $w_a$ over the sparse graph. The last two columns give the relative change in edge $F_1$ inside the anchor graph and its ratio to the change outside.}
\label{tab:ablation}
\vspace{-1.5ex}
\footnotesize
\setlength{\tabcolsep}{2.8pt}
\begin{tabular}{lccccc}
\toprule
Setting & Top-1$\uparrow$ & Edge $F_1\uparrow$ & CFSD$\downarrow$ & $\Delta F_1^{\text{in}}$ & In/Out \\
\midrule
Primary only & 0.613 & 0.606 & 0.181 & -- & -- \\
\multicolumn{6}{l}{\emph{+ Secondary pass, dose $w_a$}} \\
\quad $w_a{=}0.25$ & 0.597 & 0.624 & 0.174 & +6.2\% & 8.5 \\
\quad $w_a{=}0.50$ & 0.580 & 0.635 & 0.169 & +11.4\% & 8.3 \\
\quad $w_a{=}0.75$ & 0.545 & 0.645 & 0.164 & +15.4\% & 7.7 \\
+ CA skip & 0.637 & 0.621 & 0.182 & +1.7\% & 0.9 \\
\rowcolor{gray!20}
\textsc{AnswerPrint} & 0.607 & 0.647 & 0.169 & +13.3\% & 4.1 \\
\bottomrule
\end{tabular}
\vspace{-3ex}
\end{table}

\begin{table}[t]
\centering
\caption{\textbf{Design choices} on Reel-30. 
\emph{Top:} how the reel enters StyleID. Pooled AdaIN is our primary grade, and the \emph{per-frame} rows run once per reel frame and then  select outputs. 
\emph{Bottom:} alternative mattes for the secondary pass.}
\label{tab:design}
\vspace{-1.5ex}
\small
\setlength{\tabcolsep}{4pt}
\begin{tabular}{lccc}
\toprule
 & Top-1$\uparrow$ & Edge $F_1\uparrow$ & LPIPS$\downarrow$ \\
\midrule
\multicolumn{4}{l}{\textbf{\emph{Reel entry (StyleID)}}} \\
Central frame & $0.563$ & $0.582$ & $0.504$ \\
\rowcolor{gray!20}
Pooled AdaIN (ours) & $0.613$ & $0.606$ & $0.483$ \\
Mean K/V & $0.220$ & $0.067$ & $0.710$ \\
Concat.\ K/V & $0.485$ & $0.655$ & $0.414$ \\
Per-frame, colour pick & $0.515$ & $0.599$ & $0.501$ \\
Per-frame, CSD pick$^{\ast}$ & $0.650$ & $0.549$ & $0.521$ \\
Per-frame, pixel mean & $0.553$ & $0.597$ & $0.462$ \\
\midrule
\multicolumn{4}{l}{\textbf{\emph{Matte (matched coverage)}}} \\
\rowcolor{gray!20}
Graph (ours) & $0.607$ & $0.647$ & $0.458$ \\
Canny$^{\ast}$ & $0.597$ & $0.660$ & $0.457$ \\
Mask2Former & $0.592$ & $0.649$ & $0.458$ \\
PiDiNet & $0.592$ & $0.647$ & $0.457$ \\
Random (3 draws) & $0.594$ & $0.636$ & $0.462$ \\
\bottomrule
\end{tabular}
\\[1pt]
\begin{minipage}{\linewidth}
\centering
\footnotesize
$^{\ast}$Circular with one of the metrics (shared representation).
\end{minipage}
\vspace{-4ex}
\end{table}

\vspace{-1ex}
\section{Ablation Study}
\label{sec:ablation}

\noindent {\bf Component analysis.}
\label{sec:component}
We split edge $F_1$ into scores inside and outside the anchor graph's $8$-pixel token footprint, and report the change in each over the primary grade (\tabref{tab:ablation}): a genuine structural effect raises $F_1$ only where it acts, whereas indiscriminate copying lifts both alike. The secondary pass concentrates its gain where it operates  ($8.3{\times}$ inside vs.\ outside), and the anchor dose $w_a$ trades identity against structure, film Top-1 falling $0.613\!\to\!0.545$ as edge $F_1$ rises $0.606\!\to\!0.645$ (\appref{app:dose}); we adopt $w_a{=}0.5$. The cross-attention skip lifts both regions near-uniformly ($1.7\%$ vs.\ $1.9\%$), as expected of removing a content-independent offset, and raises film Top-1 from $0.613$ to $0.637$.

\vspace{1ex}
\noindent{\bf Which matte?} 
\label{sec:matte}
We replace only the matte by a Canny edge map~\citep{canny1986edge}, Mask2Former panoptic boundaries~\citep{cheng2022mask2former}, PiDiNet learned contours~\citep{su2021pidinet} or random tokens, each matched to the graph's token count on every image (\tabref{tab:design}, bottom). What matters is placing the matte on structure rather than which detector finds it: the graph beats random draws of equal size by $0.011$ edge $F_1$, while all structural mattes agree within $0.015$ in film Top-1. The graph matches learned and semantic mattes while being training-free and weight-free.

\begin{figure}[t]
  \centering
  \includegraphics[width=\linewidth]{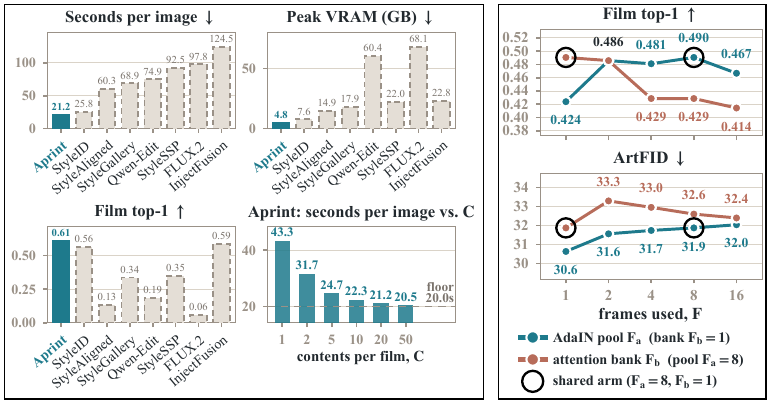}
  \vspace{-4ex}
  \caption{Ablations on cost and reel entry.}
  \label{fig:ablation}
  \vspace{-4ex}
\end{figure}

\vspace{1ex}
\noindent {\bf Cost analysis.}
\label{sec:exp-cost}
Figure~\ref{fig:ablation} (left) reports seconds per image and peak memory at Reel-30's twenty contents per film (\appref{app:baselines}). APrint needs $4.8$\,GB at peak, against $7.6$ to $22.8$\,GB for the diffusion baselines and $60.4$ to $68.1$\,GB for the editors, and takes $21.2$\,s per image: the reel's inversion is paid once per film, so the cost falls from $43.3$\,s for a single shot towards a floor of $20.0$\,s. StyleGallery's optimisation ($68.9$\,s), Qwen-Image-Edit ($74.9$\,s), StyleSSP ($92.5$\,s) and FLUX.2 ($97.8$\,s) cost three to five times as much per image.

\vspace{1ex}
\noindent{\bf Where a reel enters.}
\label{sec:where-set-enters}
A reel can enter a single-reference method at several points. In a controlled sweep over randomly drawn frames \footnote{Random rather than medoid frames isolate set size from frame selection. The sweep runs on a 210-image subset with its own held-out split.}
(\figref{fig:ablation}, right), pooling more frames into the AdaIN statistics raises film Top-1 from $0.424$ to $0.490$, whereas widening the attention bank lowers it to $0.414$: tone is distributional, so averaging over frames cancels the palette bias of any single scene, while texture is already captured from one frame and only diluted by more. As shown in \tabref{tab:design} (top), averaging keys and values blurs the bank into a washed-out mean (film Top-1 $0.220$), and concatenating them preserves structure (edge $F_1$ $0.655$) but dilutes identity ($0.485$). Running StyleID once per frame multiplies the cost by $F_a$ and still falls short, whether outputs are averaged ($0.553$) or selected by colour ($0.515$); selecting by CSD reaches $0.650$, but optimises the very embedding the probe is trained on and lowers edge $F_1$ to $0.549$. Pooled statistics is thus the only option that improves identity and structure together at no extra cost, so we pool $F_a$ frames and bank the medoid as $F_b$ (with more details in \appref{app:frameselection}).

\begin{figure}[t]
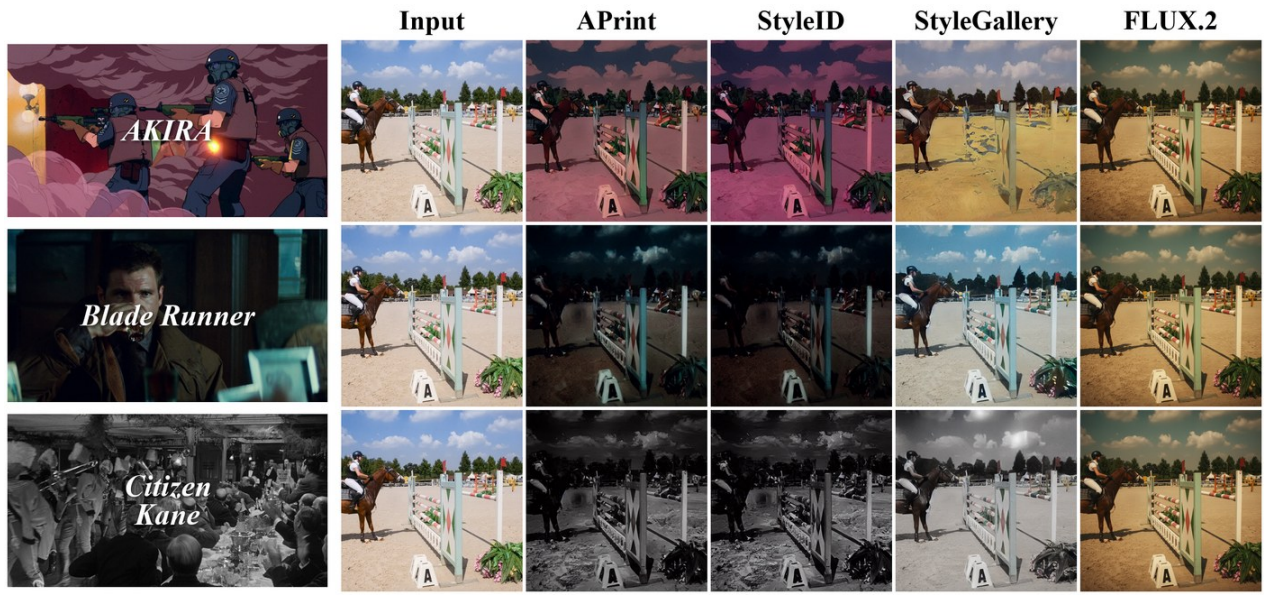

  \centering
  \animategraphics[width=\linewidth]{12}{Figure/horsejump-high/}{000}{023}
  \vspace{-4ex}
  \caption{\textbf{A Video Diagnostic.} \textbf{\emph{Plays in Adobe Acrobat/Reader.}} For more visualizations, please refer to \appref{app:video}.}
  \label{fig:video}
  \vspace{-3ex}
\end{figure}

\vspace{-1ex}
\section{Towards Practice: A Video Diagnostic}
\label{sec:exp-video}
In film post-production phase, FLT can extend naturally from stills to video.
Although APint is a still-image method, we make an initial foray into the temporal dimension: grading 8 DAVIS \cite{pont2017davis} clips\footnote{DAVIS clips span fur, plumage, a dancer, a rider, vehicles and a boat, and centre-cropped to $512{\times}512$ like the stills.} ($24$ frames each) independently toward three films in Reel-30.  APrint keeps a warping error \citep{lai2018blind} of $0.0048$ under RAFT flow \citep{teed2020raft} with occlusion masking, against $0.0024$ for the ungraded clip, $0.0056$ for StyleID and $0.0193$ for StyleGallery, and every graded frame is still assigned to its own film. Played back, APrint's clips show no visible flicker: tone, grain and painted texture hold from frame to frame and the clips read as footage from the film, whereas StyleGallery repaints its strokes and colour patches on every frame. APrint is thus preliminarily usable on short clips without any temporal module, and we leave explicit temporal modelling to future work.

%% file: section/conclusion.tex
\vspace{-1ex}
\section{Conclusion}
\label{sec:conclusion}
\vspace{-1ex}
A film's look is authored in the grading suite, and every new shot must still be matched to it. We study this step as \textit{film look transfer}, analyse what it demands of an automatic method, and introduce Reel-30 to measure film identity and local structure. AnswerPrint follows the colourist's two passes on a frozen diffusion model: a primary grade pools the reel, and a secondary pass restores the content's structure inside a sparse matte. It balances look and structure better than existing methods at a fraction of their cost.
Human studies and video trial
further validate the practical values. As \film{Casablanca} puts it,\textit{``this is the beginning of a beautiful friendship''} between the human artists and AI model.

%% file: supp_build.tex
\clearpage
\maketitlesupplementary
\appendix

\setcounter{figure}{0}
\setcounter{table}{0}
\setcounter{equation}{0}
\renewcommand{\thefigure}{A\arabic{figure}}
\renewcommand{\thetable}{A\arabic{table}}
\renewcommand{\theequation}{A\arabic{equation}}
\renewcommand{\theHfigure}{A\arabic{figure}}
\renewcommand{\theHtable}{A\arabic{table}}
\renewcommand{\theHequation}{A\arabic{equation}}
\section{Reel-30}
\label{app:reel30}

In the era of photochemical film, the spool that carried the developed footage was called a reel. We borrow the name for the set of frames drawn from a film that defines its look.

\vspace{1ex}
\noindent \textbf{A Reel is a look.} Compared with a single style image, a reel is a more meaningful reference for post-production: \eg, as shown in \figref{fig:teaser}, a colourist grading toward \film{Dune} works from the film's cool, tactile desert palette as a whole, not from one shot. A single frame is confounded by its own scene, subject and lighting, and when it enters a model as the reference the transferred look drifts with it. Averaging over several frames of the same film keeps what they share and suppresses what is idiosyncratic to any one of them, much as a colourist pins a look down more easily after seeing many shots of it.

\vspace{1ex}
\noindent \textbf{Statistics.} Reel-30 has $30$ films in eight categories, from animation to live action and from monochrome to saturated. Each film comes with a curated pool of up to $50$ frames, from which the protocol draws $F$ frames as the reel (8 frames as default; selection in \appref{app:frameselection}); the remaining frames are held out to train and test the film-identity probe (\appref{app:eval}). 
The benchmark pairs the reels with the $20$ content photographs of the standard training-free style-transfer protocol \citep{chung2024styleid}.

Table~\ref{tab:reel30-films} lists each film's \textbf{category}, its \textbf{compactness}, the mean CSD cosine of its curated frames to their own centroid (higher means the frames agree on one look and form a more reliable reel), its \textbf{stand-alone identifiability}, the recall of a linear probe that separates it from the other films ($4$-fold cross-validation; $0.86$--$1.00$ across candidates), and \textbf{the size} of its frame pool. The categories are a reporting device, not a selection criterion.

\begin{table}[t]
\centering
\caption{The thirty films of Reel-30, grouped by category. \emph{Comp.}\ is the mean CSD cosine of a film's curated frames to their centroid; \emph{ident.}\ is stand-alone identifiability; \emph{frames} is the size of the curated pool.}
\vspace{-1ex}
\label{tab:reel30-films}
\footnotesize
\setlength{\tabcolsep}{4pt}
\resizebox{\linewidth}{!}{%
\begin{tabular}{@{}rllccc@{}}
\toprule
\# & film & category & comp.\ & ident.\ & frames \\
\midrule
1  & Days of Heaven                  & arthouse        & $0.705$ & $1.00$ & $50$ \\
2  & Lawrence of Arabia              & arthouse        & $0.583$ & $0.94$ & $50$ \\
3  & Moonrise Kingdom                & arthouse        & $0.630$ & $0.92$ & $50$ \\
4  & The Fall                        & arthouse        & $0.559$ & $0.92$ & $50$ \\
5  & The Grand Budapest Hotel        & arthouse        & $0.624$ & $0.94$ & $50$ \\
6  & Suspiria                        & arthouse        & $0.581$ & $0.98$ & $50$ \\
\midrule
7  & Feeling from Mountain and Water & cn-animation    & $0.781$ & $1.00$ & $26$ \\
8  & Havoc in Heaven                 & cn-animation    & $0.708$ & $1.00$ & $26$ \\
\midrule
9  & Farewell My Concubine           & cn-liveaction   & $0.565$ & $0.94$ & $50$ \\
10 & Hero                            & cn-liveaction   & $0.523$ & $0.96$ & $50$ \\
11 & In the Mood for Love            & cn-liveaction   & $0.673$ & $0.94$ & $50$ \\
12 & Raise the Red Lantern           & cn-liveaction   & $0.688$ & $1.00$ & $50$ \\
13 & Shadow                          & cn-liveaction   & $0.764$ & $1.00$ & $50$ \\
\midrule
14 & Punch-Drunk Love                & crime-drama     & $0.628$ & $0.98$ & $50$ \\
15 & The Godfather                   & crime-drama     & $0.562$ & $0.94$ & $50$ \\
16 & The Social Network              & crime-drama     & $0.705$ & $1.00$ & $50$ \\
\midrule
17 & AKIRA                           & jp-animation    & $0.691$ & $0.90$ & $50$ \\
18 & Paprika                         & jp-animation    & $0.640$ & $0.92$ & $50$ \\
\midrule
19 & Citizen Kane                    & monochrome      & $0.748$ & $0.98$ & $50$ \\
\midrule
20 & Blade Runner (1982)             & sci-fi           & $0.618$ & $0.96$ & $50$ \\
21 & Dune                            & sci-fi           & $0.558$ & $0.96$ & $50$ \\
22 & Pacific Rim                     & sci-fi           & $0.628$ & $1.00$ & $50$ \\
23 & The Matrix                      & sci-fi           & $0.727$ & $0.98$ & $50$ \\
24 & Transformers                    & sci-fi           & $0.657$ & $0.93$ & $30$ \\
\midrule
25 & Coraline                        & west-animation  & $0.553$ & $0.90$ & $50$ \\
26 & Isle of Dogs                    & west-animation  & $0.520$ & $0.94$ & $50$ \\
27 & Klaus                           & west-animation  & $0.733$ & $1.00$ & $50$ \\
28 & Loving Vincent                  & west-animation  & $0.794$ & $1.00$ & $50$ \\
29 & Spider-Man: Into the Spider-Verse & west-animation & $0.614$ & $1.00$ & $50$ \\
30 & The Triplets of Belleville      & west-animation  & $0.738$ & $1.00$ & $50$ \\
\bottomrule
\end{tabular}}
\end{table}

\subsection{Film Choice}
\label{sec:selection}
To test whether a model captures a specific film's style rather than a generic cinematic grade, we select films with pronounced and mutually distinct visual signatures. The roster therefore avoids redundancy: no two films should be near-duplicates in look. Each film is represented by the normalised mean CSD embedding of its frames (\appref{app:eval}), and from $38$ candidates we keep the $30$-film subset that minimises the largest centroid cosine between any two kept films, \ie, $\tau^{\star}{=}0.611$. The eight dropped films each break a redundant pair above $\tau^\star$ (Table~\ref{tab:reel30-redundant}). We keep the most self-consistent one (highest mean compactness), where several subsets reach this bound.

\begin{table}[t]
\centering
\caption{The ten redundant film pairs above $\tau^\star$, ranked by centroid cosine in the CSD embedding.}
\vspace{-1ex}
\label{tab:reel30-redundant}
\small
\setlength{\tabcolsep}{5pt}
\begin{tabular}{@{}llc@{}}
\toprule
film A & film B & centroid cosine \\
\midrule
Howl's Moving Castle & Spirited Away         & $0.926$ \\
AKIRA                & Ghost in the Shell    & $0.792$ \\
Citizen Kane         & Schindler's List      & $0.743$ \\
Paprika              & Spirited Away         & $0.699$ \\
Blade Runner         & Seven                 & $0.691$ \\
Blade Runner         & Joker                 & $0.686$ \\
Howl's Moving Castle & Paprika               & $0.674$ \\
Seven                & The Social Network    & $0.669$ \\
Arcane               & Klaus                 & $0.654$ \\
Puss in Boots        & Into the Spider-Verse & $0.650$ \\
\bottomrule
\end{tabular}
\end{table}

\subsection{Evaluation protocol}
\label{app:eval}

\noindent \textbf{CSD embedding.} Our film-identity measures for compactness, redundancy and film Top-1 share one frozen CSD encoder \citep{somepalli2024csd}: the public \texttt{CSD-ViT-L} checkpoint, an OpenAI CLIP ViT-L/14 backbone loaded through \texttt{open\_clip} with the visual projection disabled, so that the $1024$-dimensional pooled feature passes through the model's style head to a $768$-dimensional style embedding, $L_2$-normalised at output. Inputs follow the standard CLIP evaluation transform at $224{\times}224$, and letterbox bars are removed from film frames.

\vspace{1ex}
\noindent \textbf{Film Top-1.} A multinomial logistic-regression probe over these embeddings, trained only on real held-out frames. To report its ceiling we take a $75/25$ stratified split of the real frames, giving $0.987$ against a chance level of $0.033$; the probe is then refit on all held-out frames and applied once to the outputs. Film Top-1 is the fraction of outputs it assigns to their source film, and category Top-1 counts an assignment to any film of the same category as correct.

\vspace{1ex}
\noindent \textbf{Style similarity.} The mean pairwise cosine similarity between an output and the held-out frames of its source film, following the DINO protocol of DreamBooth \citep{ruiz2023dreambooth} with DINOv2 features.

\vspace{1ex}
\noindent \textbf{Edge $F_1$.}
We detect Canny edges \citep{canny1986edge} on the greyscale output and content (read as $512{\times}512$ 8-bit), and match them with a $2$-pixel tolerance implemented by dilation with a $5{\times}5$ square: \emph{precision} is the fraction of output edge pixels lying in the dilated content edges, \emph{recall} the fraction of content edge pixels lying in the dilated output edges, and $F_1=2PR/(P{+}R)$. Scores are computed per image and averaged over the outputs. For the component analysis (\secref{sec:component}) we score the same quantity separately \emph{inside} the region the anchor graph acts on and \emph{outside} it.

\vspace{1ex}
\noindent \textbf{CFSD.} As a second structure measure we report the Content Feature Structural Distance of StyleID \citep{chung2024styleid}, which depends only on how image patches correlate with each other, not on their colour. Let $\mF\in\R^{hw\times c}$ be the \texttt{conv3} feature map of VGG19 for an image; we form $\mM=\mF\mF^{\top}$, turn each row into a distribution $\mS=[\sm(\mM_i)]_{i=1}^{hw}$, and take the KL divergence between the content's and the output's maps,
\begin{equation}
\mathrm{CFSD}=\frac{1}{hw}\sum_{i=1}^{hw} D_{\mathrm{KL}}\!\big(\mS^{c}_i \,\big\|\, \mS^{cs}_i\big).
\label{eq:cfsd}
\end{equation}

\vspace{1ex}
\noindent \textbf{ArtFID family.} Following the standard protocol we also report FID \citep{heusel2017fid} between outputs and reference frames, LPIPS \citep{zhang2018lpips} between output and content, and ArtFID \citep{wright2022artfid} $=(1{+}\text{LPIPS})(1{+}\text{FID})$. How these measures relate to human judgment is analysed in \appref{app:metrics}.

\subsection{Human studies}
\label{app:human}

\noindent \textbf{Blind preference.}

\noindent \emph{Raters.} Eight volunteers recruited through our partner film academy: seven film practitioners and one cinematography student, with $0$ to $13$ years in the field (median $2$). Participation was voluntary, and every rater gave informed consent.

\vspace{1ex}
\noindent \emph{Items.} $24$ film--content pairs from $24$ distinct films, stratified by category. APrint is compared with ten baselines spanning all method groups: StyleID, InjectFusion, InstantStyle and StyleSSP (single-reference), Ruta \etal\ and StyleGallery (multi-reference), MKL and IDT (Color Transfer), and Qwen-Image-Edit and FLUX.2 (editing); each baseline is judged on twelve of the pairs. All outputs are the Reel-30 outputs scored in \tabref{tab:main}, centre-cropped to $512{\times}512$.

\vspace{1ex}
\noindent \emph{Presentation.} An offline page showed four held-out frames of the film, the raw photograph, and the two outputs as A and B, with method names hidden. Two questions were answered separately per trial, forced choice without ties:
\begin{quote}
    \textit{\emph{(i)} which output better matches the film's look and is the more usable grade; \\
    \emph{(ii)} which better keeps the photograph's structure.}
\end{quote}

\vspace{1ex}
\noindent \emph{Allocation.} Each rater judged all $120$ trials ($8$ judgments per trial and $960$ per question). Trial order and the side on which APrint appeared were randomised per rater, with APrint on the left in half the trials. Twelve trials were shown a second time later in the session with the sides swapped, to measure self-consistency. The median time per trial was $7.3$\,s.

\vspace{1ex}
\noindent \emph{Analysis.} For each baseline we report the \textbf{\emph{preference rate}}, \ie, the share of judgments that chose APrint, with a $95\%$ Wilson interval \citep{wilson1927interval} for its uncertainty (\tabref{tab:human}). We test whether each rate differs from chance ($0.5$) with a two-sided exact binomial test, Holm-corrected across the ten baselines. For the rate pooled over all baselines we also give a $95\%$ interval from a bootstrap that resamples raters and items independently ($10^4$ draws). \textbf{\emph{Inter-rater agreement}} is Fleiss' $\kappa$ \citep{fleiss1971kappa} per question and \textbf{\emph{self-consistency}} is the share of repeated trials on which a rater made the same choice again. Whether our metrics agree with these judgments is analysed in \appref{app:metrics}.

\begin{table}[t]
\centering
\caption{\textbf{Blind 2AFC study.} Share of judgments preferring APrint over each baseline, with $95\%$ Wilson intervals; $96$ judgments per baseline ($8$ raters $\times$ $12$ pairs). Rates below $50\%$ mean the baseline was preferred. 
$^{*}$ two-sided exact binomial test, $p<0.05$ after Holm correction across the ten baselines.}
\label{tab:human}
\small
\setlength{\tabcolsep}{3pt}
\resizebox{\linewidth}{!}{%
\begin{tabular}{lcc}
\toprule
APrint vs. & Look & Structure \\
\midrule
MKL & $50.0\%$ $[40.2, 59.8]$ & $22.9\%$ $[15.6, 32.3]^{*}$ \\
IDT & $60.4\%$ $[50.4, 69.6]$ & $20.8\%$ $[13.9, 30.0]^{*}$ \\
InstantStyle & $50.0\%$ $[40.2, 59.8]$ & $65.6\%$ $[55.7, 74.4]^{*}$ \\
InjectFusion & $91.7\%$ $[84.4, 95.7]^{*}$ & $99.0\%$ $[94.3, 99.8]^{*}$ \\
StyleID & $67.7\%$ $[57.8, 76.2]^{*}$ & $71.9\%$ $[62.2, 79.9]^{*}$ \\
StyleSSP & $45.8\%$ $[36.2, 55.8]$ & $52.1\%$ $[42.2, 61.8]$ \\
StyleGallery & $47.9\%$ $[38.2, 57.8]$ & $33.3\%$ $[24.7, 43.2]^{*}$ \\
Ruta \etal & $81.2\%$ $[72.3, 87.8]^{*}$ & $94.8\%$ $[88.4, 97.8]^{*}$ \\
Qwen-Image-Edit & $77.1\%$ $[67.7, 84.4]^{*}$ & $59.4\%$ $[49.4, 68.7]$ \\
FLUX.2 & $75.0\%$ $[65.5, 82.6]^{*}$ & $22.9\%$ $[15.6, 32.3]^{*}$ \\
\midrule
Pooled ($n{=}960$) & $64.7\%$ $[61.6, 67.6]^{*}$ & $54.3\%$ $[51.1, 57.4]^{*}$ \\
\quad rater$\times$item bootstrap & $[57.8, 71.5]$ & $[46.6, 61.8]$ \\
Fleiss' $\kappa$ & $0.21$ & $0.45$ \\
Self-consistency & $73\%$ & $77\%$ \\
\bottomrule
\end{tabular}}
\end{table}

\vspace{1ex}
\noindent \textbf{Manual grading.} 

\begin{figure*}[t]
  \centering
  \includegraphics[width=\textwidth]{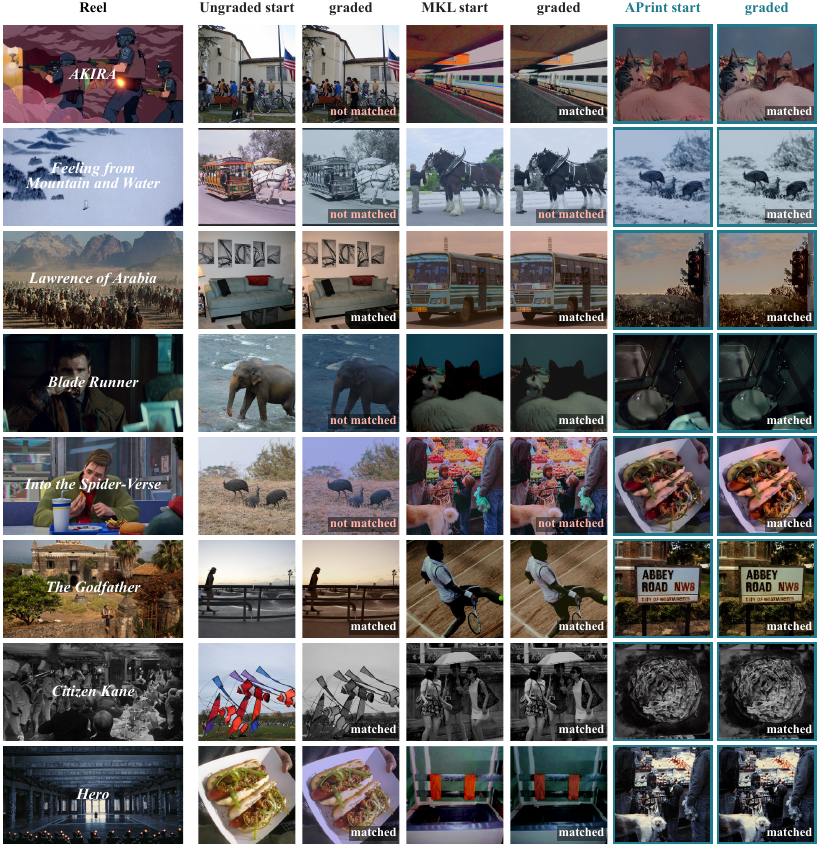}
  \vspace{-3ex}
  \caption{\textbf{Manual grading, all eight films.} One row per film; from left to right, a reel frame, then each start the colourist received (the ungraded plate, MKL and APrint) followed by the grade reached from it, marked \emph{matched} or \emph{not matched}. Each start is a different content of the same film. Matching is a partly subjective professional judgment, so these counts indicate a tendency rather than a measured rate.}
  \label{fig:grading_all}
\end{figure*}

\noindent One professional colourist graded $8$ Reel-30 films, one per category, each from three starts: the ungraded plate, the output of MKL and the output of APrint. Each start used a different content of the film, and the $24$ tasks ran in a seeded order with the starts interleaved. The colourist received each start with the film's drawn reel frames, film names hidden, and graded it in \emph{DaVinci Resolve} from scratch until it matched the reel to their professional standard, or stopped when it could not be matched; a second colourist, who did not see the start, confirmed each match. Figure~\ref{fig:grading_all} shows every start and the grade reached from it.

APrint starts were matched for all $8$ films, MKL starts for $6$ and plates for $4$. The plate could not be matched for \film{AKIRA}, \film{Feeling from Mountain and Water}, \film{Into the Spider-Verse} and \film{Blade Runner}, and MKL for \film{Feeling from Mountain and Water} and \film{Into the Spider-Verse}. Grading changes colour and tone but cannot redraw a photograph, so a look that lies in its rendering is out of reach from a plate or a colour transfer, whereas an APrint start already carries it and leaves the colourist only local refinements of its colour.
The colourist also changed APrint starts the least: averaged over pixels and channels, the absolute difference 
between start and grade ($0$--$255$) has a median of $8.6$ for APrint, $12.3$ for MKL and $18.7$ for the plate.

\section{Main results}
\label{app:results}

\begin{figure*}[t]
    \centering    \includegraphics[width=\textwidth]{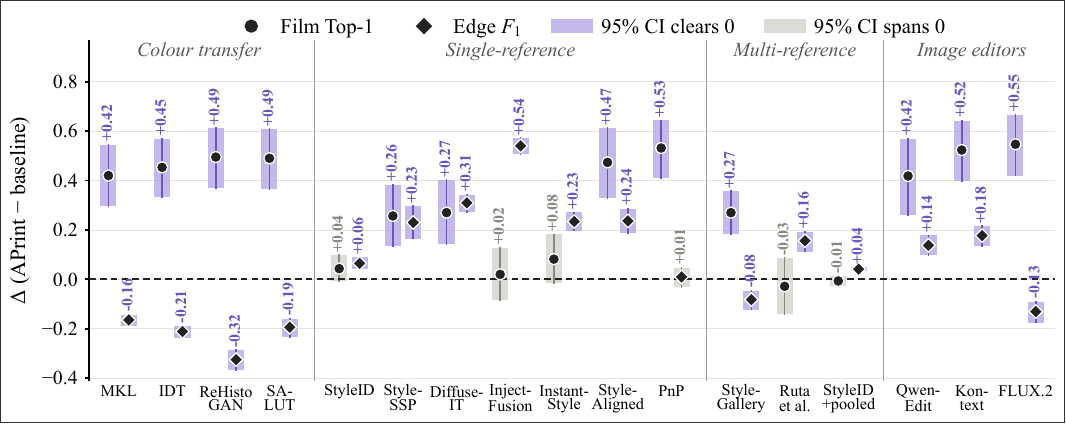}
    \vspace{-3ex}
    \caption{Film-level paired bootstrap ($10{,}000$ resamples over $30$ films) of APrint minus each baseline.}
    \label{fig:clean}
    \vspace{-1ex}
\end{figure*}

Fig.~\ref{fig:clean} tests each comparison of \tabref{tab:main} with a film-level paired bootstrap of APrint minus the baseline.

\subsection{Text-conditioned methods}
\label{app:text}
\textbf{Plug-and-Play }\cite{tumanyan2023pnp} injects features under a target prompt and takes no image to define style, so a film can reach it only through text. To place it on Reel-30 without changing what defines a style, we caption the same central frame it would otherwise receive. Captions come from Qwen2.5-VL-3B-Instruct with deterministic image-to-sentence mapping, asking only for the visual style:
\begin{quote}
\small
\textit{``Describe only the visual style of this film still: its colour palette, contrast, lighting, film grain and texture. Do not name any object, person or place. Answer with one short phrase beginning with the word `style'.''}
\end{quote}
This yields $30$ descriptions of $8$ to $36$ words (median $24$), for instance for \film{AKIRA}: \textit{\small``style The film still exhibits a gritty, dark colour palette with high contrast, dramatic lighting, and a grainy texture that evokes a sense of tension and urgency''}. Through this channel PnP barely alters the content.

\noindent Two image-conditioned baselines also need a prompt, which for them describes the content. \textbf{InstantStyle} \cite{wang2024instantstyle} receives the content caption with the quality suffix and negative prompt, and takes the film through the IP-Adapter's CLIP image embedding of the central frame. \textbf{StyleAligned} \cite{hertz2024stylealigned} generates the reference and the target in one batch, so its reference row is prompted with a caption of what is \emph{in} the film still (\eg {\small ``Man drinking at bar''} for \film{Blade Runner}) and its target row with the content caption, while the film's look reaches the target through shared attention. Content captions come from BLIP-2 and are identical across films.

\subsection{Degenerate outputs}
\label{app:mislead}
\textbf{Mislead Metrics.} A method that barely alters the target leads every content-fidelity column. Colour transfer does exactly this: ReHistoGAN has the best edge $F_1$ ($0.972$), LPIPS ($0.141$) and ArtFID ($24.27$) in \tabref{tab:main}, and IDT the best CFSD ($0.038$), yet none of the four colour methods carries the film (film Top-1 at most $0.187$). PnP shows the same pattern among diffusion methods: it edits almost nothing (film Top-1 $0.075$ against a chance of $0.033$) and ranks near the top on LPIPS and ArtFID. In the other direction, StyleAligned records the best FID ($17.92$) while transplanting the reference into the output so heavily that little of the content survives. A single read-out can thus be won by a degenerate output in either direction, which is why we read identity and structure together, as a trade-off.

\vspace{1ex}
\noindent \textbf{Visual Distortions.} Beyond the columns of \tabref{tab:main}, the outputs of \figref{fig:sota} fail in four recurring ways, each invisible to at least one metric.
\emph{Imposed rendering}: StyleSSP turns live-action plates into an animated surface, smoothing skin, merging the trusses of a bridge into one mass and cleaning a cluttered water surface, and Ruta \etal\ lay heavy painterly strokes over \film{Blade Runner} plates as over \film{Loving Vincent}.
\emph{Reference content leaking in}: InjectFusion replaces parts of the plate with objects and rooms from the reference frames, and Qwen-Image-Edit returns a mosaic of reel frames in place of the target.
\emph{Identity drift}: DiffuseIT, InstantStyle and Ruta \etal\ redraw faces until the person is no longer recognisable.
\emph{Washed-out output}: StyleAligned dissolves the subject into a pale haze.
None of these transfers the film's look, since each erases or replaces the content rather than grading it.

\section{Method details}
\label{app:hyper}

\subsection{Frame selection and sensitivity}
\label{app:frameselection}

\textbf{The draw.} For each film, $F$ frames are drawn at random from its pool with a fixed per-film seed. The frames are then ordered by distance to their colour centre,
\begin{equation}
f^{\star}=\argmin_{f\in\mathcal{S}}\big\|\,\hat{\vd}^{f}-\bar{\hat{\vd}}\,\big\|_2 ,
\label{eq:central}
\end{equation}
where $\hat{\vd}^{f}$ is the $9$-D global colour descriptor of frame $f$ (channel means, log channel standard deviations and the three cross-channel correlations), standardised over the drawn frames, and $\bar{\hat{\vd}}$ is their mean.

In Sec.~\ref{sec:alight}, APrint pools $F_a$ of the $F$ drawn frames into its AdaIN statistics (all of them by default, $F_a{=}F{=}8$) and banks the most central one ($F_b{=}1$), whereas single-reference baselines receive the same central frame, and multi-reference baselines receive all $F$ frames. The rule needs only the supplied frames, never the full film.

\vspace{1ex}
\noindent \textbf{Frame sensitivity.} To measure how much a single-reference method depends on the chosen still, we run our base model, StyleID, once per centrality rank on Reel-30. Film Top-1 falls from $0.563$ at the most central frame to $0.387$ at the least central one (mean $0.456\pm0.056$; Table~\ref{tab:frame-sensitivity}). Even within one film, a single frame conflates the film's look with its own scene, lighting and framing. We therefore give every single-reference method the most central frame, so no baseline gains from a luckier still.

\begin{table*}[t]
\centering
\caption{Frame sensitivity on Reel-30: StyleID with each of the eight drawn frames as its reference, ordered by colour centrality.}
\vspace{-1ex}
\label{tab:frame-sensitivity}
\small
\setlength{\tabcolsep}{6pt}
\begin{tabular}{@{}lccccc@{}}
\toprule
Style reference & film Top-1 $\uparrow$ & category $\uparrow$ & style sim.\ $\uparrow$ & edge $F_1$ $\uparrow$ & ArtFID $\downarrow$ \\
\midrule
rank 0 \emph{(most central)}$^{\ddagger}$ & $0.563$ & $0.905$ & $0.213$ & $0.582$ & $30.65$ \\
rank 1 & $0.530$ & $0.878$ & $0.196$ & $0.588$ & $30.62$ \\
rank 2 & $0.447$ & $0.845$ & $0.200$ & $0.576$ & $32.24$ \\
rank 3 & $0.435$ & $0.882$ & $0.189$ & $0.542$ & $30.90$ \\
rank 4 & $0.443$ & $0.892$ & $0.199$ & $0.593$ & $30.93$ \\
rank 5 & $0.442$ & $0.882$ & $0.204$ & $0.590$ & $30.24$ \\
rank 6 & $0.403$ & $0.848$ & $0.168$ & $0.593$ & $29.60$ \\
rank 7 \emph{(least central)} & $0.387$ & $0.857$ & $0.159$ & $0.438$ & $29.39$ \\
\midrule
mean $\pm$ sd over the eight & $0.456 \pm 0.056$ & $0.874 \pm 0.020$ & $0.191 \pm 0.017$ & $0.563 \pm 0.050$ & $30.57 \pm 0.83$ \\
\midrule
\textsc{APrint} (the reel) & $\mathbf{0.607}$ & $\mathbf{0.920}$ & $\mathbf{0.222}$ & $\mathbf{0.647}$ & $30.77$ \\
\bottomrule
\end{tabular}
\\[2pt]
{\footnotesize $^{\ddagger}$ The frame that single-reference baselines receive and APrint banks ($F_b$) in \tabref{tab:main}.}
\vspace{-1ex}
\end{table*}

\vspace{-1ex}
\subsection{Anchor dose}
\label{app:dose}
The anchor graph is built by Alg.~\ref{alg:anchor} (\appref{app:graph}). The dose $w_a$ trades identity against local structure monotonically (\tabref{tab:anchor-dose}): from no anchoring through $w_a{=}0.25$ to $w_a{=}0.75$, film Top-1 falls $0.613\to0.597\to0.580\to0.545$ while edge $F_1$ rises $0.606\to0.624\to0.635\to0.645$. 

\vspace{1ex}
\noindent \textbf{Curve variants.} The straight-line graph is nearly empty on organic content such as animals or portraits, which offer few long lines and sharp corners. A curved contour layer reaches these objects, but at a clear cost in film identity (\tabref{tab:anchor-dose}, $cw_a$ rows). We keep the straight graph at $w_a{=}0.5$ as the default and record the curved layer as an ablation.

\begin{table}[t]
\centering
\caption{Anchor dose. $w_a$ blends the straight anchor graph back at the anchor tokens; $cw_a$ the same graph with a curved contour layer added. Best in each column in \textbf{bold}.}
\vspace{-1ex}
\label{tab:anchor-dose}
\scriptsize
\setlength{\tabcolsep}{2.5pt}
\resizebox{\linewidth}{!}{%
\begin{tabular}{@{}lccccc@{}}
\toprule
anchor & film T1 $\uparrow$ & sty.\ $\uparrow$ & edge $F_1$ $\uparrow$ & ArtFID $\downarrow$ & CFSD $\downarrow$ \\
\midrule
No anchor & $\mathbf{0.613}$ & $\mathbf{0.226}$ & $0.606$ & $31.80$ & $0.181$ \\
\midrule
$w_a = 0.25$ & $0.597$ & $0.222$ & $0.624$ & $31.62$ & $0.174$ \\
\rowcolor{gray!20}
$w_a = 0.50$ & $0.580$ & $0.218$ & $0.635$ & $31.41$ & $0.169$ \\
$w_a = 0.75$ & $0.545$ & $0.212$ & $0.645$ & $31.13$ & $0.164$ \\
\midrule
$cw_a = 0.25$ & $0.553$ & $0.209$ & $0.644$ & $31.32$ & $0.161$ \\
$cw_a = 0.50$ & $0.452$ & $0.186$ & $0.673$ & $31.07$ & $0.145$ \\
$cw_a = 0.75$ & $0.377$ & $0.164$ & $\mathbf{0.695}$ & $\mathbf{30.74}$ & $\mathbf{0.135}$ \\
\bottomrule
\end{tabular}}
\vspace{-1ex}
\end{table}

\vspace{-1ex}
\section{Metrics and human judgment}
\label{app:metrics}
\vspace{-0.5ex}
We compare our metrics with the raters of the blind study (\appref{app:human}) at two levels.\textbf{ Across baselines,} ArtFID tracks the pairwise preference for APrint over the ten baselines (Spearman $\rho{=}0.83$ on the mean of the two questions), and the agreement comes from its content term: LPIPS alone reaches $\rho{=}0.91$, whereas FID alone reaches $\rho{=}{-}0.32$. Distribution distance to the reference frames therefore does not measure film look, which is why we score identity with the film probe and read identity and structure as two separate axes. 
\textbf{Trial by trial,} we ask whether our metrics pick the same output as the raters. The raters' choice on a trial is the output chosen by the majority of the eight, while trials split $4{:}4$ have no majority and are left out. The metric's choice, for structure, is the output with the higher edge $F_1$, and for look, the output to which the film-identity probe of film Top-1 gives the higher probability of the source film. Agreement is the fraction of eligible trials on which metric and raters choose the same output. As a reference we also report the always-APrint agreement. For structure, edge $F_1$ agrees on $75.2\%$ of $109$ eligible trials ($95\%$ Wilson $[66.4, 82.4]$; Cohen's $\kappa{=}0.49$; always-APrint $55.0\%$). For look, the probe agrees on $65.1\%$ of $106$ trials ($[55.6, 73.5]$; $\kappa{=}0.16$; always-APrint $70.8\%$). The disagreements concentrate on two baselines: raters chose APrint on all $12$ such trials against InjectFusion and all $11$ against Ruta \etal, whereas the probe prefers the baseline on $6$ of each, outputs that carry strong film identity but break the plate.

\section{Process correction}
\label{app:process}

\subsection{Skipping cross-attention}
\label{app:Clean~Gate}
\begin{figure}[t]
    \centering
    \includegraphics[width=0.46\textwidth]{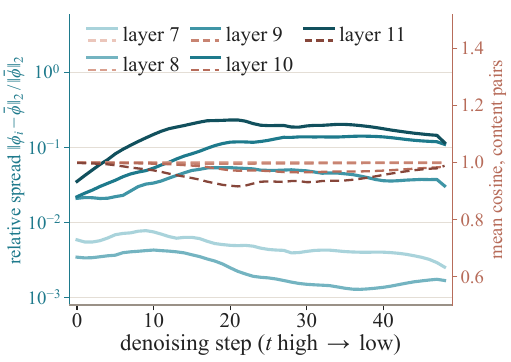}
    \vspace{-1ex}
    \caption{The CA output varies little in norm (blue) and stays near-constant in direction (red).}
    \label{fig:ca_skip}
    \vspace{-1ex}
\end{figure}

Alongside the self-attention modified in \eqref{eq:attn}, each decoder block contains a cross-attention (CA) module. In the unconditional regime we adopt, the prompt is empty, so its keys and values are formed from a fixed set of padding embeddings shared across all inputs and timesteps, and the only input-dependent term is the query. Measured across contents (Figure~\ref{fig:ca_skip}), the CA output is dominated by a shared, content-independent direction: the mean cosine between per-content outputs stays above $0.95$ at every layer and step, and approaches $1.0$ in the layers where our operator acts. CA therefore contributes the same offset to every image, carrying neither the content nor the film. We skip it in the decoder blocks that receive the reel, leaving the self-attention that carries the film's look untouched. Since the offset is content-independent, the deletion is expected to act uniformly over the image, which \secref{sec:component} confirms.

\section{Baselines}
\label{app:baselines}

Unless noted otherwise, every baseline runs with its published code at their recommended settings; below we give the details needed to reproduce our runs. Each is run on the Reel-30 pairs and receives the same $512{\times}512$ centre crops of the drawn frames. Outputs at other resolutions are resized to $512$ with Lanczos before scoring. Text-conditioned methods are described in \appref{app:text}.

\subsection{Colour transfer methods}
All are fitted to the reel without diffusion. \textbf{MKL} \citep{pitie2007mkl} is the closed-form Monge--Kantorovich linear map between the content's RGB distribution and the distribution pooled over the film frames (letterbox removed, frames subsampled to $384$ pixels). \textbf{IDT }\citep{pitie2007automated} is iterative distribution transfer over the fixed rotations followed by its re-graining step, with the same pooled pixels of the reel.
\textbf{ReHistoGAN} \citep{afifi2021histogan} uses its universal model, with the mean RGB-uv histogram of the reel as target, which runs at $256$ pixels with pyramid upsampling to $512$. \textbf{SA-LUT} \citep{gong2025salut} uses two-stage inference (network at $512$, 4D LUT applied at full resolution), and since it takes one reference, it receives the central frame.

\subsection{Multi-reference methods}
\textbf{Ruta \etal} \citep{ruta2025multistyle} is our reimplementation, since no code is public. Following the paper: SD1.5, ControlNet depth and line-art from the content; an IP-Adapter projection fine-tuned on the drawn frames (${\sim}100$ steps) and averaged over them; DDIM inversion of the frames with their self-attention values k-means clustered per layer, timestep and head to one image's token count, keeping the nearest real key--value pairs; and an average-style image generated from the averaged embedding, whose query/key and latent statistics drive AdaIN at every self-attention layer and step. We set ControlNet scales at $1.0$, IP-Adapter scale at $1.0$, AdamW at $10^{-4}$, $10$ Lloyd iterations, CFG $7.5$, $50$ DDIM steps and seed $0$.

\vspace{1ex}
\noindent \textbf{StyleID with the reel.} We also run StyleID over all drawn frames, letting the reel enter in several ways: pooling the AdaIN statistics while keeping the central frame's attention (our primary grade), averaging keys and values token-wise or concatenating them across frames, or running each frame separately and combining the outputs by CSD selection (the output nearest the centroid of the reference frames), by colour selection, or by averaging pixels (ablation in Sec.~\ref{sec:where-set-enters}).

\subsection{Image editors}
The editors take one to three images, so each receives the reel as one sheet of the drawn frames, with an instruction that names no film.
\textbf{Qwen-Image-Edit} \citep{wu2025qwenimage} and \textbf{FLUX.2 [dev]} \citep{bfl2025flux2} take the target and the sheet ($512$-pixel tiles) as two input images and output $1024{\times}1024$, at their model-card settings (Qwen-Image-Edit: $40$ steps, true CFG $4.0$; FLUX.2: $50$ steps, guidance $4.0$; both seed $0$, bf16), with the instruction
\begin{quote}
\small
\emph{``Re-grade image 1 so that it has the cinematic look of images 2: the same colour grading, tonality, contrast, lighting mood and film texture. Keep the content, composition, people, objects and geometry of image 1 exactly unchanged.''}
\end{quote}
\textbf{FLUX.1 Kontext [dev]} \citep{labs2025kontext} accepts one input image, so the target and the sheet ($256$-pixel tiles) share one $1536{\times}512$ canvas and the left $512{\times}512$ of the output is kept, at its model-card settings ($28$ steps, guidance $2.5$, seed $0$), with the same instruction phrased for panels:
\begin{quote}
\small
\emph{``Two panels. Re-grade the LEFT photograph so it has the cinematic look of the RIGHT film frames: the same colour grading, tonality, contrast, lighting mood and film texture. Keep the left photograph's content, composition, people, objects and geometry exactly unchanged, and keep the two-panel layout.''}
\end{quote}

\noindent \textbf{Content check.} For every output we take the Pearson correlation of a $64{\times}64$ greyscale copy with the target and with each reel frame. Outputs below $0.3$ correlation with the target number $0$ for APrint, $9$ for FLUX.2, $40$ for Kontext and $145$ for Qwen-Image-Edit, out of $600$; in a random sample of the last, every one is a re-arranged mosaic of reel frames.

\subsection{Cost analysis}
All the methods run alone on one NVIDIA GH200 (120\,GB) at the settings above, on one film (\film{AKIRA}), first for one content ($T_1$) and then for eleven ($T_{11}$). Peak memory is the maximum of \texttt{nvidia-smi} over the run. The marginal cost per image is $m{=}(T_{11}{-}T_1)/10$ and the one-off cost, model loading plus any per-film preparation such as inversion or fine-tuning, is $T_1{-}m$. We report $(T_1{-}m)/20+m$, the cost per image at twenty contents per film.

\section{Anchor graph}
\label{app:graph}
The anchor graph is constructed from the content alone, once before sampling, by Alg.~\ref{alg:anchor}: line segments are detected and merged into polylines, corners are kept only near a retained segment, and small connected components are discarded.

\begin{algorithm}[t]
\LinesNotNumbered
\SetAlgoLined
\DontPrintSemicolon
\SetKwInOut{Input}{input}\SetKwInOut{Output}{output}
\SetKwFunction{Merge}{Merge}\SetKwFunction{LSD}{LSD}
\SetKwFunction{Corners}{Corners}\SetKwFunction{Comps}{Components}
\SetKwFunction{Raster}{Raster}
\Input{content image $I$; layers $\mathcal{L}$; dose $w_a$}
\Output{anchored output $\phi^{cs}$ at every $\ell$}
\BlankLine
$g \leftarrow \mathrm{grey}(I)$\;
$S \leftarrow \Merge\big(\LSD(g);\ \alpha_{\mathrm{tol}},\gamma,\delta_\perp\big)$\;
$S \leftarrow$ the $N_\ell$ longest of $\{s : \lVert s\rVert \ge \ell_{\min}\}$\;
$V \leftarrow \{p \in \Corners(g;N_c) : \min_{s} \operatorname{dist}(p,s) \le \rho\}$\;
$\mathcal{G} \leftarrow S \cup V$\;
$\mathcal{G} \leftarrow \mathcal{G} \setminus \{k \in \Comps(\mathcal{G}) : \operatorname{extent}(k) < e_{\min}\}$\;
$R \leftarrow \Raster(\mathcal{G})$\;
\ForEach{$\ell \in \mathcal{L}$}{
$M_\ell \leftarrow$ reduce $R$ onto $\ell$'s grid by \eqref{eq:maxpool}\;
$\phi^{cs} \leftarrow$ blend $\phi^{cs}$ with $\phi^{c}$ at $M_\ell$ by \eqref{eq:anchor}\;
}
\caption{Construction of the anchor graph. Reported configuration: $N_c{=}80$, $\ell_{\min}{=}64$, $N_\ell{=}20$, $\rho{=}6$, $e_{\min}{=}48$, $w_a{=}0.5$; \textsc{Merge} tolerances $\alpha_{\mathrm{tol}}{=}10^\circ$, $\gamma{=}16$, $\delta_\perp{=}4$.}
\label{alg:anchor}
\end{algorithm}

\vspace{1ex}
\noindent \textbf{Coverage.} On the twenty contents the graph lights on average $8.96\%$ of tokens at the finer injected grid ($64{\times}64$) and $16.6\%$ at the coarser one ($32{\times}32$), varying by image because it anchors long structure, which a building offers far more of than an animal. The operator is thus a sparse intervention on both grids, 
which keeps the inside and outside regions of \secref{sec:component} separable.

\vspace{1ex}
\noindent \textbf{Matched-coverage alternatives.} In the matte comparison of \tabref{tab:design} (bottom), each alternative ranks the $64{\times}64$ tokens by its own per-token response and lights, on every image, as many tokens as the graph does. At the $32{\times}32$ grid, where neighbouring tokens merge, Canny then lights $17.9\%$, PiDiNet $17.7\%$ and semantic boundaries $19.2\%$, against the graph's $16.6\%$, while random selection is matched at both grids. The random row is the mean of three independent draws, which span film Top-1 $0.592$--$0.598$, edge $F_1$ $0.635$--$0.636$ and ArtFID $31.27$--$31.34$.

\section{Video diagnostic}
\label{app:video}
\textbf{Setup.} Eight DAVIS clips \citep{pont2017davis} (bear, blackswan, boat, car-roundabout, dance-twirl, dog, horsejump-high, train), $24$ frames each at $512{\times}512$, are graded frame by frame, with no temporal model, toward three reels of contrasting look: \film{AKIRA}, \film{Blade Runner} and \film{Citizen Kane}. Every method treats each frame as an independent still with the same settings as on Reel-30.

\begin{table}[t]
\centering
\caption{Warping error ($\times10^{-3}$, lower is steadier) of frame-wise grading on eight DAVIS clips. The ungraded clips score $2.4$.}
\vspace{-1ex}
\label{tab:video}
\small
\setlength{\tabcolsep}{4pt}
\begin{tabular}{@{}lcccc@{}}
\toprule
& \film{AKIRA} & \film{Blade Runner} & \film{Citizen Kane} & mean \\
\midrule
StyleGallery & $16.9$ & $22.7$ & $18.5$ & $19.3$ \\
FLUX.2 & $8.8$ & $9.8$ & $9.2$ & $9.3$ \\
StyleID & $4.0$ & $1.6$ & $11.1$ & $5.6$ \\
\rowcolor{gray!15}
\textsc{APrint} & $4.8$ & $2.3$ & $7.3$ & $4.8$ \\
\bottomrule
\end{tabular}
\end{table}

\vspace{1ex}
\noindent \textbf{Warping error.} For consecutive frames we estimate optical flow on the ungraded clip with RAFT \citep{teed2020raft}, warp output frame $t{+}1$ back onto frame $t$, and average the squared colour difference over pixels that pass a forward--backward occlusion check \citep{lai2018blind}. Table~\ref{tab:video} reports the mean over clips for each film. APrint's error is twice that of the ungraded clip and below FLUX.2's on $23$ of $24$ clip--film pairs and StyleGallery's on all of them. StyleID, which shares our backbone and edits less, is lower on \film{AKIRA} and \film{Blade Runner} and higher on \film{Citizen Kane}. Every APrint frame is assigned to its own film by the probe.

\section{More visualizations}
\label{app:visual}
\begin{figure*}[t]
  \centering
  \includegraphics[width=\textwidth]{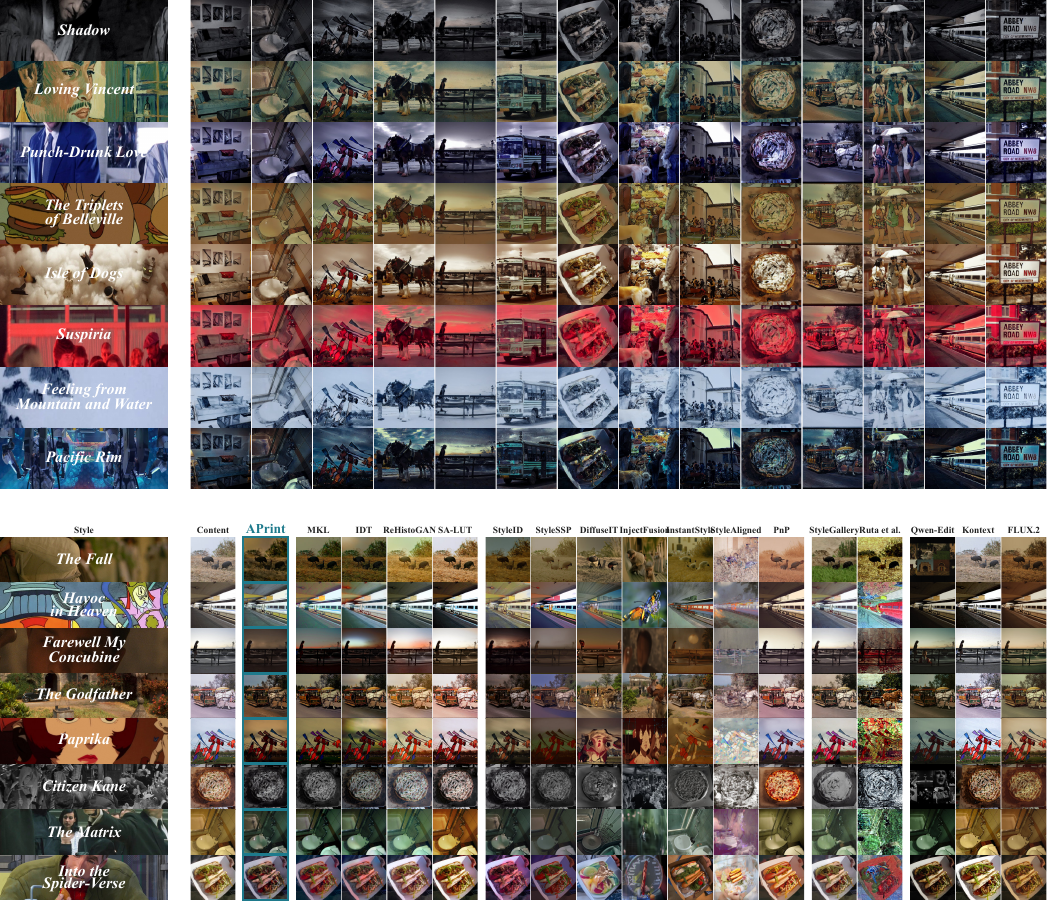}
  \vspace{-3ex}
  \caption{\textbf{More results on Reel-30} (zoom in for detail). \emph{Top:} APrint across eight films and fourteen contents. \emph{Bottom:} comparison with every method of \tabref{tab:main} on eight further films; the style column shows each reel's central frame.}
  \label{fig:vis}
\end{figure*}

Figure~\ref{fig:results} shows APrint's outputs across films and contents, and \figref{fig:vis} gives a larger gallery (top) and extends the comparison of \figref{fig:sota} to eight further films (bottom). Please zoom in for detail.

\section{Limitations}
\label{app:limits}
\begin{figure}[t]
  \centering
  \includegraphics[width=\linewidth]{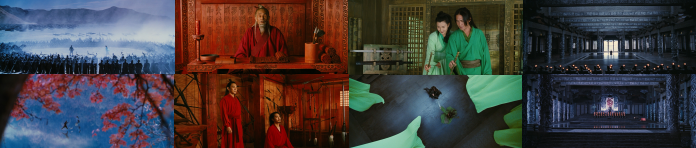}
  \caption{
  Frames from the Reel-30 pool of \film{Hero}, which span four disjoint colour regimes.
  }
  \label{fig:hero}
  \vspace{-2ex}
\end{figure}

Our approach has several limitations that also mark directions for future work.
\begin{itemize}[itemsep=2pt, topsep=2pt]
 \item \textbf{A frozen backbone caps the ceiling.} We keep the diffusion backbone frozen and add no weights, which makes the method light and training-free but bounds it by the generator's own capacity: content the backbone renders poorly, such as fine high-frequency texture, high resolutions or domains under-represented in its prior, cannot be recovered by any of our passes.
 \item \textbf{One look per reel.} The reel-level formulation assumes a film shares a consistent tone and texture across frames. Some films narrate through deliberately disjoint colour regimes (\figref{fig:hero}), or are driven by story rather than a distinctive visual signature, so no single look exists to transfer; on such inputs the task itself is under-defined.
 \item \textbf{The matte favours rigid structure.} The line--corner graph covers long, straight structure well and organic subjects poorly (\appref{app:graph}), so faces, fur and foliage receive the least restoration, whereas a curved contour layer reaches them at a cost in film identity (\appref{app:dose}).
 \item \textbf{A still-image method.} We recover the static tone and texture a still-image method can reach and add no temporal model. Graded frame by frame, short clips stay steady (\appref{app:video}), but this is not guaranteed for long shots or fast motion, and evaluation uses photographs rather than graded camera originals. Cinematic language beyond the look, including rhythm, camera motion and narrative structure, lies outside this scope.
\end{itemize}

\begin{quote}
\itshape
\film{``It's easier to make a colour look good than it is for colour to service a story.''}
\par\vspace{0.4em}
\hspace*{\fill}\film{--- Roger Deakins}
\end{quote}

\subsection*{Ethics statement}

\textbf{Data provenance and copyright.} Our new benchmark is built from publicity and gallery stills of thirty copyrighted motion pictures, obtained through lawful access and used solely for non-commercial academic research. Copyright in the original audiovisual works remains with the respective rights holders. We redistribute no film frames and no video. Our release contains only a manifest, together with the film and frame selection procedures, the evaluation code, and the preprocessing scripts needed to reproduce the benchmark; readers obtain the stills themselves through lawful channels and can verify by digest that they are running on our exact inputs.

\noindent \textbf{Human studies.} All raters and colourists took part voluntarily and gave informed consent. For each rater we recorded only their role, years in the field and display type, and no participant data is released. No formal ethics review was carried out because the studies were minimal-risk tasks that involved only judging or grading images, and collected no identifying or sensitive personal data.

\begin{figure*}[t]
  \centering
  \includegraphics[width=\textwidth]{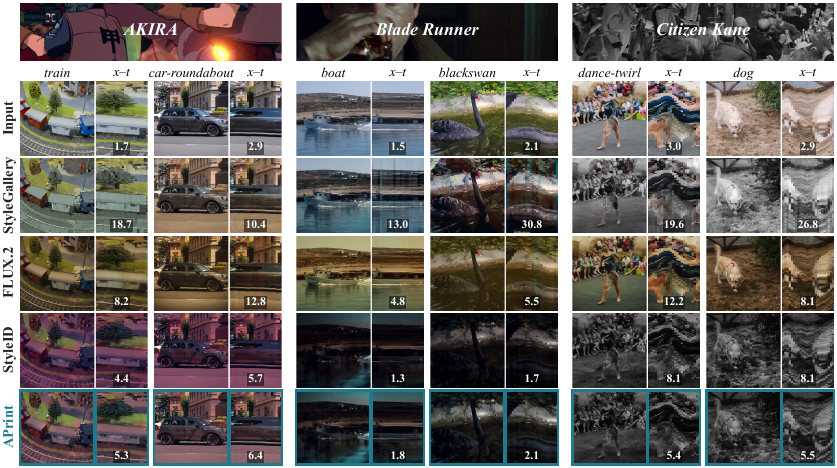}
  \vspace{-3ex}
  \caption{\textbf{Frame-wise grading of DAVIS clips.} Two clips per reel, each shown by its middle frame and an $x$--$t$ slice (the centre pixel column over all $24$ frames), where flicker would appear as vertical streaks and numbers give the warping error ($\times10^{-3}$).}
  \label{fig:video_diag}
\end{figure*}